\documentclass[lettersize,journal]{IEEEtran}
\usepackage{amsmath,amsfonts}
\usepackage{algorithmic}
\usepackage{algorithm}
\usepackage{array}
\usepackage[caption=false,font=normalsize,labelfont=sf,textfont=sf]{subfig}
\usepackage{textcomp}
\usepackage{stfloats}
\usepackage{url}
\usepackage{verbatim}
\usepackage{graphicx}
\usepackage{cite}
\usepackage{times}
\usepackage{epsfig}
\usepackage{amssymb}
\usepackage{gensymb}
\usepackage{multirow}
\usepackage{color}
\usepackage{caption2}
\usepackage{array}
\usepackage{booktabs}
\usepackage{threeparttable}
\usepackage{pifont}
\usepackage[dvipsnames]{xcolor}
\usepackage[colorlinks=true,linkcolor=OrangeRed,citecolor=green,urlcolor=magenta,]{hyperref}

\newcommand{\cmark}{\ding{51}}%
\newcommand{\xmark}{\ding{55}}%
\definecolor{orange}{rgb}{1,0.5,0}

\begin{document}

\title{Blind VQA on 360{\textdegree} Video via Progressively Learning from Pixels, Frames and Video}

\author{
	Li Yang,~\IEEEmembership{Graduate Student Member,~IEEE}, Mai Xu,~\IEEEmembership{Senior Member,~IEEE}, Shengxi Li,~\IEEEmembership{Member,~IEEE}, Yichen Guo,~\IEEEmembership{Graduate Student Member,~IEEE}, Zulin Wang,~\IEEEmembership{Member,~IEEE} 
\thanks{This work was supported by the NSFC projects 61876013, 61922009 and 61573037.(Corresponding author: Mai Xu.)}
\thanks{L. Yang, M. Xu, S. Li, Y. Guo and Z. Wang are with the School of Electronic and Information Engineering, Beihang University, Beijing 100191, China (e-mail: \{LiYang2018; MaiXu; 16711024\}@buaa.edu.cn, ShengxiLi2014@gmail.com).}}

\markboth{Journal of \LaTeX\ Class Files,~Vol.~14, No.~8, August~2021}%
{Shell \MakeLowercase{\textit{et al.}}: A Sample Article Using IEEEtran.cls for IEEE Journals}


\maketitle

\begin{abstract}
Blind visual quality assessment (BVQA) on 360{\textdegree} video plays a key role in optimizing immersive multimedia systems. 
When assessing the quality of 360{\textdegree} video, human tends to perceive its quality degradation from the viewport-based spatial distortion of each spherical frame to motion artifact across adjacent frames, ending with the video-level quality score, \textit{i.e.}, a progressive quality assessment paradigm. However, the existing BVQA approaches for 360{\textdegree} video neglect this paradigm. In this paper, we take into account the progressive paradigm of human perception towards spherical video quality, and thus propose a novel BVQA approach (namely ProVQA) for 360{\textdegree} video via progressively learning from pixels, frames and video. Corresponding to the progressive learning of pixels, frames and video, three sub-nets are designed in our ProVQA approach, \textit{i.e.}, the spherical perception aware quality prediction (SPAQ), motion perception aware quality prediction (MPAQ) and multi-frame temporal non-local (MFTN) sub-nets. The SPAQ sub-net first models the spatial quality degradation based on spherical  perception  mechanism of human. Then, by exploiting motion cues across adjacent frames, the MPAQ sub-net properly incorporates motion contextual information for quality assessment on 360{\textdegree} video. Finally, the MFTN sub-net aggregates multi-frame quality degradation to yield the final quality score, via exploring long-term quality correlation from multiple frames. The experiments validate that our approach significantly advances the state-of-the-art BVQA performance on 360{\textdegree} video over two datasets, the code of which has been public in \url{https://github.com/yanglixiaoshen/ProVQA.}
\end{abstract}


\section{\textbf{Introduction}}
\IEEEPARstart{V}{irtual} reality (VR), as a new type of immersive technology, tends to be increasingly popular in a variety of fields, such as automotive industry, healthcare and entertainment, \textit{etc}. As an essential type of VR content, 360{\textdegree} videos have been flooding into human daily life. Different from traditional 2D videos, 360{\textdegree} videos can provide users with an original visual experience, benefiting from the immersive surrounding and interactive pattern in VR. To meet the quality of experience (QoE) of users in 360$\times$180{\textdegree} viewing range, the resolution of 360{\textdegree} videos is extraordinarily high, \textit{e.g.}, 8K, 16K or even higher, which may cause dramatically heavy burdens on the storage and transmission of 360{\textdegree} videos \cite{xu2020state, Deng2021LAU-Net}. Consequently, to alleviate such burdens, it is inevitable to compress 360{\textdegree} videos for saving bit-rates and bandwidth \cite{mathias2019standardization}. However, heavy compression on 360{\textdegree} videos may introduce ``VR sickness", \textit{e.g.}, dizziness, disorientation and nausea, which awfully degrades the subjective QoE of viewers. To measure the compression performance, visual quality assessment (VQA) for 360{\textdegree} videos is urgently demanded to evaluate the quality degradation caused by compression \cite{xu2020state, LIU201876, Guo2021A}, and further guide the optimization of VR system for better QoE.

In past decades, despite prominent progress on VQA for 2D videos \cite{li2019quality,mittal2015completely, korhonen2019two, dendi2020no, chen2020rirnet, saad2014blind, liu2018end, manasa2016optical, wu2019quality, tu2021ugc, sinno2018large}, comparatively few efforts have been devoted to VQA for 360{\textdegree} videos \cite{zhang2018subjective, zhang2017subjective, singla2017comparison, xu2018assessing, de2020a, li2019viewport, li2018bridge, xu2020viewport, chai2021blind}. Specifically, the works on VQA for 360{\textdegree} videos can be classified into two types, \textit{i.e.}, subjective and objective VQA. In subjective VQA, subjects are required to rate quality scores for the viewed 360{\textdegree} videos under specialized VR environment, and thus the collected scores can reflect the realistic and natural subjective opinions of perceptual quality. Nevertheless, the subjective VQA suffers from heavy labor-consuming,  making it impractical for real-time monitoring the quality of 360{\textdegree} videos in practical applications. Therefore, it is crucial to study the objective VQA for 360{\textdegree} videos, which is capable of automatically predicting the subjective quality of the input 360{\textdegree} videos. Recently, some works of full reference VQA (FR VQA) for 360{\textdegree} videos have been proposed \cite{xu2018assessing, de2020a, li2019viewport, li2018bridge, xu2020viewport}. For example, Xu \textit{et al.} \cite{xu2020viewport} proposed a viewport-based convolutional neural network (V-CNN) approach on VQA for 360{\textdegree} videos, which has a multi-task architecture composed of a viewport proposal network (VP-net) and viewport quality network (VQ-net). The VP-net handles the auxiliary tasks of camera motion detection and viewport proposal, while the VQ-net accomplishes
the auxiliary task of viewport saliency prediction and the main task of VQA. Compared with FR VQA, blind VQA (BVQA), also called no reference VQA (NR VQA), has potentially much broader applicability, since they can predict the quality in the absence of the reference 360{\textdegree} videos. Moreover, the dilemma of acquiring the reference 360{\textdegree} videos in the real-world circumstance greatly promotes the demand of BVQA for 360{\textdegree} videos.

To our best knowledge, there exists only one approach towards BVQA for 360{\textdegree} videos \cite{chai2021blind}. Based on the two-stream architecture, NR-OVQA \cite{chai2021blind} was proposed to fine-tune the pre-trained deep bilinear convolutional neural network (DB-CNN) \cite{zhang2018blind} and 3D neural network (C3D) model \cite{tran2015learning}, for extracting intra- and inter-frame quality features of the input 360{\textdegree} video, respectively. The final quality score can be obtained by aggregating the quality features from two streams. Unfortunately, NR-OVQA neglects the progressive perception mechanism of human in assessing the quality of 360{\textdegree} video, \textit{i.e.}, perceiving from pixels, frames and video. In this regard, the general BVQA framework should accord with human perception towards video quality, and take the spatial-temporal regularities of 360{\textdegree} video into consideration. In specific, it can be summarized as three key factors for BVQA on 360{\textdegree} video, \textit{i.e.}, spherical visual characteristics, motion artifacts between adjacent frames and long-term correlation across multiple inconsecutive frames. To take the above factors into account, we propose a novel approach for BVQA on 360{\textdegree} video, via progressively learning from pixels, frames and video, which is called ProVQA. 

Specifically, the network of our ProVQA approach includes spherical perception aware quality prediction (SPAQ) subnet, motion perception aware quality prediction (MPAQ) sub-net and multi-frame temporal non-local (MFTN) sub-net. We first propose the SPAQ sub-net to model the spatial quality degradation conditioned on spherical perception mechanism. In this way, the pixel-wise quality map for each 360{\textdegree} frame can be obtained. To exploit the motion cues from adjacent frames, the MPAQ sub-net is developed to further refine the pixel-wise quality map as the frame-level quality map. It allows to incorporate more quality-related temporal context in the predicted quality map. Finally, we design the MFTN sub-net for aggregating multiple frame-level maps of all 360{\textdegree} frames to yield the overall quality score, such that the long-term quality correlation among multiple inconsecutive frames can be captured for BVQA on 360{\textdegree} video. Finally, the extensive experimental results show the effectiveness of our ProVQA approach in BVQA on  360{\textdegree} video.
       
In summary, the contributions of our paper are three-fold:
\begin{itemize}
	\item We propose a novel ProVQA framework for BVQA on 360{\textdegree} video, which progressively learns from pixels, frames and video to model and assess the quality degradation of 360{\textdegree} video. 
	\item We develop three sub-nets for our ProVQA approach, which learn the spatial quality degradation from pixels, the motion artifacts from frames, and video-level quality aggregation from 360{\textdegree} video, respectively. 
	\item We conduct extensive experiments to show that our approach outperforms 13 state-of-the-art approaches for BVQA on 360{\textdegree} video by large margins in both efficiency and efficacy. 	
\end{itemize}

\section{\textbf{Related works}}

\subsection{BVQA for 360{\textdegree} image and video}

Recently, there have emerged several works in the domain of BVQA for 360{\textdegree} image over the years \cite{zhou2021omnidirectional, xu2020blind, truong2019non, jiang2021cubemap, sun2019mc360iqa, Yang2021Spatial}. Generally speaking, most of these works use CMP or viewport images as the network input, instead of the whole 360{\textdegree} image, with the purpose of avoiding geometric distortion caused by sphere-to-plane projection. Additionally, most works take into account human perception in designing the networks, for better performance of quality prediction. Truong \textit{et al.} \cite{truong2019non} proposed a CNN-based model on BVQA for 360{\textdegree} image, where multiple patches are cropped from the 360{\textdegree} image and allocated with different weights according the equator-bias mechanism \cite{rai2017dataset} before inputting the CNN model. Then, all weighed patches are flowed into the CNN model for predicting quality score. Jiang \textit{et al.} \cite{jiang2021cubemap} designed a cubemap-based perception-driven BVQA framework for 360{\textdegree} image, where the cubemap quality features and attention distortion features represent global and local quality, respectively. Moreover, Xu \textit{et al.} \cite{xu2020blind} proposed to leverage Graph Convolution Network (GCN) in BVQA, where the interaction between different viewports is deeply mined for local quality estimation. The global quality features, which are learned based on DB-CNN, are integrated with the local quality features to regress onto the final perceptual quality.

In comparison to 360{\textdegree} images, to the best of our knowledge, there is only one work for on BVQA for 360{\textdegree} video, which is called NR-OVQA \cite{chai2021blind}. In \cite{chai2021blind}, the spherical 360{\textdegree} video is first projected onto six equal-area 2D videos via cubemap projection (CMP), which are viewed as the input of CNN. Then, a two-stream CNN model was developed to extract the intra-frame and inter-frame information for modeling quality degradation, where the pr-trained DB-CNN and C3D models are utilized for modeling the spatial and temporal quality features, respectively. However, NR-OVQA fails to take into account the crucial factors in BVQA for 360{\textdegree} video, \textit{i.e.}, spherical characteristics, motion perception and long-term temporal correlation, and this significantly degrades their performance on quality assessment. In this paper, we embed these three factors in our ProVQA approach to quantify the quality degradation, thus obtaining superior performance in BVQA for 360{\textdegree} videos. 

\begin{figure*}[!tb]
	\begin{center}
		\includegraphics[width=.96\linewidth]{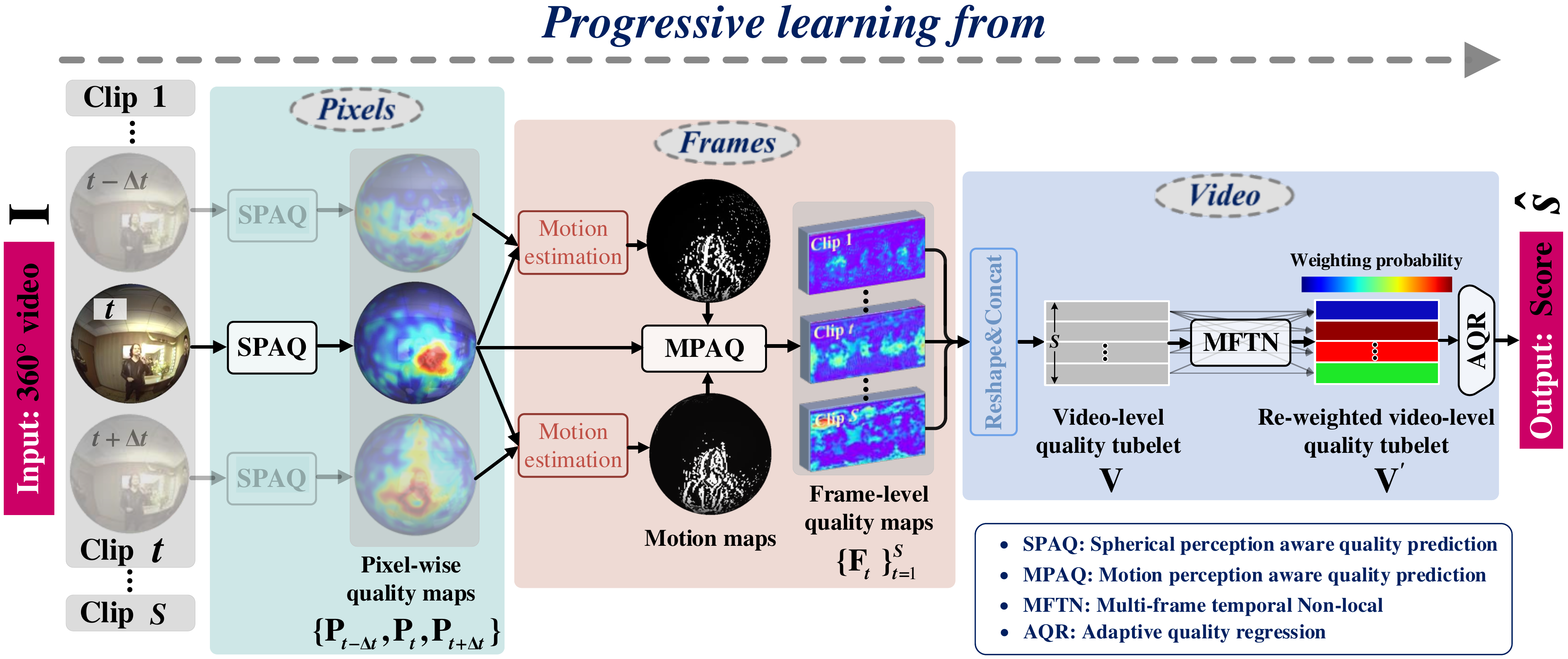}
	\end{center}
	\vspace{-0.6em}
	\caption{Framework of our proposed ProVQA approach, which includes the SPAQ, MPAQ and MFTN  sub-nets. By progressively learning from the pixels, frames and video, we can accomplish the BVQA task on 360{\textdegree} video.}
	\vspace{-0.8em}
	\label{framework}
\end{figure*}

\vspace{-0.9em}
\subsection{BVQA for 2D image and video}

For 2D image, extensive BVQA works have been proposed during the past decade, which are mainly categorized into two classes: the natural scene statistics (NSS)-based approaches \cite{mittal2012no, zhang2015feature, min2016blind, min2018blind, min2017blind}, and deep learning-based approaches \cite{yang2019sgdnet, pan2018blind, zhu2020metaiqa, fang2020perceptual, su2020blindly, chiu2020assessing, gu2020giqa, wang2021troubleshooting, ma2017dipiq, zhang2018blind}. Mittal \textit{et al.} \cite{mittal2012no} proposed a NSS-based distortion-generic BVQA model. This model does not compute distortion-specific features, such as ringing, blur, or blocking. Instead, it uses the NSS features of locally normalized luminance coefficients to quantify the possible losses of “naturalness” in images, due to the presence of distortions. Min \textit{et al.} \cite{min2017blind} utilized a new “reference” called pseudo-reference image (PRI) and developed a PRI-based BVQA framework. By calculating the structure similarity between the distorted image and its PRI, the quality score can be yielded. Yang \textit{et al.} \cite{yang2019sgdnet} proposed an end-to-end saliency-guided deep neural network (SGDNet) for 2D image BVQA. SGDNet is built on an end-to-end multi-task learning framework, in which two sub-tasks including visual saliency prediction and image quality prediction, are jointly optimized with a shared feature extractor. Regarding the challenges of BVQA on the real-world images, Su \textit{et al.} \cite{su2020blindly} developed a self-adaptive hyper network architecture composed of three sub-nets: content understanding, perception rule learning and quality predicting networks. The existing 2D image BVQA approaches are unfeasible in handling our task, since they neither have the distortion quantification in temporal dimension nor consider the geometric distortion occurring in the sphere-to-plane projection.

For 2D video, BVQA has also been extensively studied and surveyed \cite{li2019quality, mittal2015completely, korhonen2019two, dendi2020no, chen2020rirnet, saad2014blind, liu2018end, li2016spatiotemporal, xu2014no, tu2021ugc, wang2021rich, li2015no}. Basically, these works can be classified into two categories: natural video statistics (NVS)-based approaches and deep learning-based approaches. For NVS-based models \cite{mittal2015completely, li2016spatiotemporal, saad2014blind, dendi2020no, korhonen2019two}, the main efforts have been devoted to devise rational NVS features in the spatial-temporal domain for quality assessment. Saad \textit{et al.} \cite{saad2014blind} proposed a BVQA approach, in which the NVS features refer to spatial information by conducting discrete cosine transform and temporal information by calculating motion coherency. Dendi \textit{et al.} \cite{dendi2020no} utilized the mean subtracted and contrast normalized (MSCN) coefficients to quantify the distortion in 2D videos, and developed an asymmetric generalized Gaussian distribution (AGGD) to model the statistics of MSCN coefficients. Finally, the quality score can be obtained via support vector regression (SVR). However, the above NVS-based models mainly rely on hand-crafted features which are still in infancy in revealing the subjective quality of videos. To avoid such a disadvantage of hand-crafted features, many deep learning approaches \cite{liu2018end, tu2021ugc, chen2020rirnet, li2019quality} have been proposed to automatically learn spatial-temporal features for BVQA, by means of the common techniques of recurrent neural networks (RNNs) \cite{zaremba2014recurrent} or C3D. Specifically, Li \textit{et al.} \cite{li2019quality} took into account two factors, \textit{i.e.}, content-dependency and temporal-memory effects, and then developed a CNN and gated recurrent unit (GRU) model for BVQA. By fusing motion information derived from different temporal frequencies, Chen \textit{et al.} \cite{chen2020rirnet} proposed a hierarchical recurrent modeling scheme to quantify the temporal motion effect in quality assessment for 2D video. Although the features extracted by deep learning approaches possess inherent description of spatial-temporal distortion on 2D videos, it may lead to the failure in assessing the quality of 360{\textdegree} videos with unique spherical characteristics. More importantly, most 2D BVQA approaches do not progressively assess the quality of  pixel, frame and video, which are important in determining the perceptual quality of videos.

\section{\textbf{The Proposed Approach}}

In this section, we propose ProVQA as a novel BVQA approach on 360{\textdegree} videos. First, we formulate the problem of BVQA and then present the framework of our ProVQA approach. Subsequently, the primary components of our ProVQA approach are described in details. Finally, the protocol of training the ProVQA model is discussed.

\begin{figure*}[!tb]
	\begin{center}
		\includegraphics[width=1\linewidth]{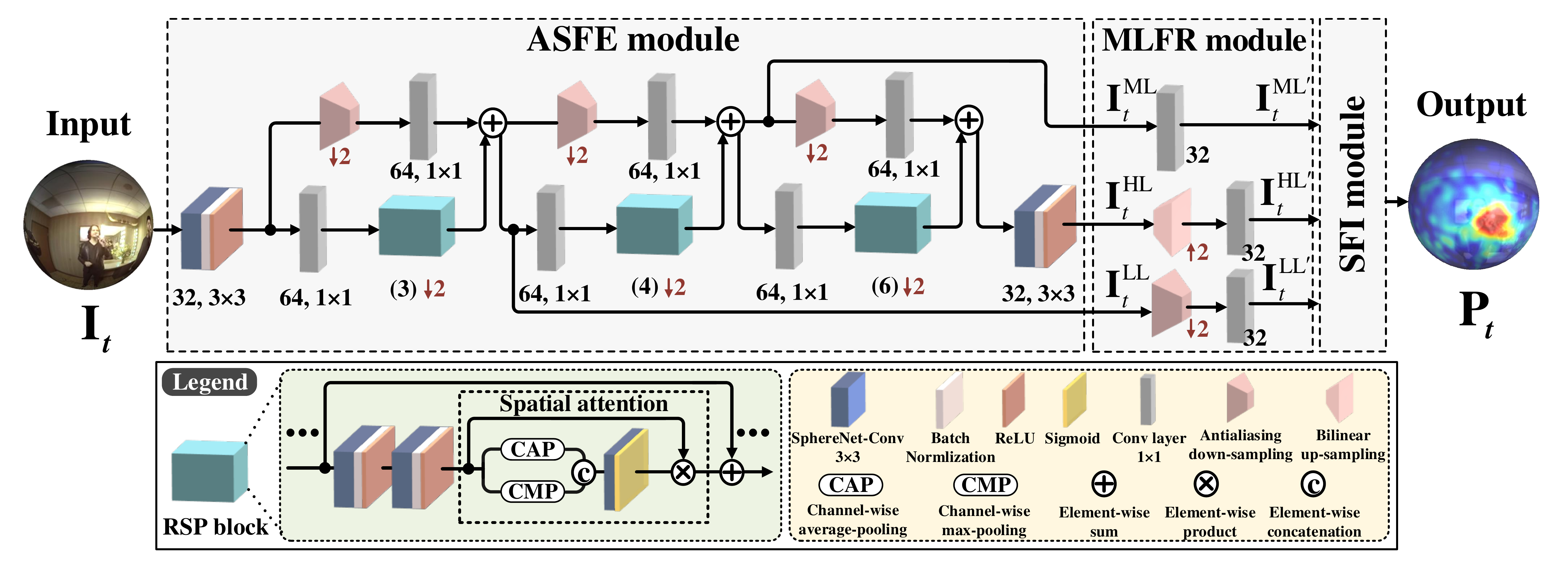}
	\end{center}
	\vspace{-0.6em}
	\caption{Architecture of our SPAQ sub-net for modeling the spatial quality degradation on 360{\textdegree} video. It consists of three novel modules, \textit{i.e.}, attention-based spherical feature extraction, multi-level feature re-scaling and selective feature integration module. For simplicity, the above modules are marked as ASFE, MLFR and SFI module in this figure. Here, ``32, 3$\times$3" denote that the number of output channels is 32 and kernel size is 3$\times$3 of the corresponding layer, respectively; ``(3)" denotes the number of RSP block; ``$\downarrow$2" and ``$\uparrow$2" denote down-sampling and up-sampling on the feature map with strides 2, respectively. }
	\vspace{-0.8em}
	\label{SPAQ}
\end{figure*}

\vspace{-1.2em}
\subsection{Framework Overview}

Assume that an impaired 360{\textdegree} video is $\mathbf{I}=\{\mathbf{I}_t\}_{t=1}^T$, where $\mathbf{I}_t$ is the $t$-th frame of the 360{\textdegree} video and $T$ is the total number of frames. Given $\mathbf{I}=\{\mathbf{I}_t\}_{t=1}^T$ as input, the ultimate goal of our ProVQA approach is to predict the subjective quality score $s$ of the impaired 360{\textdegree} video. As illustrated in Figure \ref{framework}, the framework of ProVQA is composed of three progressive stages towards the task of BVQA, \textit{i.e.}, learning quality-related features from pixels, frames and video. These progressive stages are accomplished through the sub-nets of SPAQ, MPAQ and MFTN, respectively. Generally speaking, the ProVQA approach takes the impaired 360{\textdegree} video $\mathbf{F}$ as input into the network, and outputs the corresponding predicted quality score $\hat{s}$, with the following workflow.

\begin{itemize}
	\item \textbf{Learn from pixels}. First, we sample $S$ (0 $<S \le$ $T$-2) video clips from the 360{\textdegree} video $\mathbf{I}$. Take the $t$-th clip as an example. Each clip consists of three frames, \textit{i.e.}, frame $\textbf{I}_t$ and its adjacent frames $\textbf{I}_{t-\Delta t}$ and $\textbf{I}_{t+\Delta t}$. The clip, \textit{i.e.}, $\{\textbf{I}_{t-\Delta t}, \textbf{I}_t, \textbf{I}_{t+\Delta t}\}$, is fed through the SPAQ sub-net to model the spatial quality degradation, and then we can obtain the corresponding pixel-wise quality maps $\{\textbf{P}_{t-\Delta t}, \textbf{P}_t, \textbf{P}_{t+\Delta t}\}$.  
	
	\item \textbf{Learn from frames}. Subsequently, the pixel-wise quality maps $\{\textbf{P}_{t-\Delta t}, \textbf{P}_t, \textbf{P}_{t+\Delta t}\}$ are fed into the MPAQ sub-net to generate the frame-level quality map $\textbf{F}_t$, where $\textbf{P}_{t-\Delta t}$ and $\textbf{P}_{t+\Delta t}$ serve as the supporting frames for $\textbf{P}_t$ to capture the motion in feature level. It enables the frame-level quality map to incorporate motion contextual information for BOVA on 360{\textdegree} video. 
	\item \textbf{Learn from video}. Finally, the frame-level quality maps $\{\textbf{F}_t\}_{t=1}^{S}$ from all clips are aggregated as the video-level quality tubelet $\textbf{V}$ and flowed into the MFTN sub-net, to produce the re-weighted video-level quality tubelet $\textbf{V}^{'}$. It ensures the network to explore the quality correlation among multiple inconsecutive frames. By adaptively pooling  $\textbf{V}^{'}$ via the AQR module, the overall quality score $\hat{s}$ can be obtained for the impaired 360{\textdegree} video $\mathbf{I}$.   
\end{itemize} 

More details about the structure of our ProVQA approach are discussed in the following.

\vspace{-1em}
\subsection{SPAQ sub-net}
The SPAQ sub-net mainly focuses on assessing the spatial quality degradation for each frame of 360{\textdegree} video, conditioned on spherical perception mechanism. At frame $t$, we take the impaired clip $\{\textbf{I}_{t-\Delta t}, \textbf{I}_t, \textbf{I}_{t+\Delta t}\}$ as input into the SPAQ sub-net, to yield the corresponding pixel-wise quality maps $\{\textbf{P}_{t-\Delta t}, \textbf{P}_t, \textbf{P}_{t+\Delta t}\}$. The pixel-wise quality map indicates the quality degradation in the spatial domain for each impaired frame. The architecture of the SPAQ sub-net is illustrated in Figure \ref{SPAQ}, which consists of three modules: attention-based spherical feature extraction module, multi-level feature re-scaling module and selective feature integration module.    

\textbf{Attention-based spherical feature extraction module}. Due to the prominent capacity of feature learning, ResNet \cite{he2016deep} has been widely applied in vision tasks, including BVQA on 360{\textdegree} image \cite{Yang2021Spatial, sun2019mc360iqa, xu2020blind}. However, these approaches only employ ResNet as the network backbone for feature extraction, and do not take the spherical properties into consideration. To account for the geometric properties of 360{\textdegree} video, we propose to embed the convolution layer of SphereNet \cite{coors2018spherenet} into the residual block of ResNet. This way, the distortion introduced by sphere-to-plane projection can be eliminated during feature extraction on 360{\textdegree} video, which cannot be avoided upon the standard convolution in original residual block. Furthermore, drawing upon the strong correlation between the visual attention and perceptual quality on 360{\textdegree} video \cite{xu2020viewport}, we integrate the attention mechanism into the residual block, as attention-based quality estimation in the spatial domain. Based on the above two aspects, the proposed residual spherical perception modeling (RSP) block performs as the basic operating unit, the structure of which can be seen in Figure \ref{SPAQ}. With a skip connection, each RSP block consists of two ``SConv-BN-ReLU" layers and a spatial attention operation, where SConv and BN denote the convolution layer of SphereNet and batch normalization, respectively. Here, our spatial attention operation is improved from \cite{Woo2018CBAM} by substituting the standard convolution layer with SConv. In addition to the short skip connection inside each RSP block, the long skip connection is properly added across different feature levels to reduce the training burden. Different from \cite{he2016deep}, the operations of down-sampling and 1$\times$1 convolution are employed in the long skip connection for dimension reduction and feature integration. Here, the number of RSP blocks in each level is set to 3, 4 and 6 respectively. Finally, taking the impaired frame $\textbf{I}_{t}$ as input, the features from different levels $\textbf{I}_t^{\rm LL}$, $\textbf{I}_t^{\rm ML}$ and $\textbf{I}_t^{\rm HL}$ are extracted and then flowed into the multi-level feature re-scaling module, where LL, ML and HL stand for low-level, middle-level and high-level, respectively.

\textbf{Multi-level feature re-scaling module}. Given the multi-level perceptual characteristics of human visual system (HVS) \cite{adelson1984pmi}, VQA is significantly affected by the local details and global composition of the impaired stimuli. Previous VQA works \cite{Yang2017An, Yan2020Blind, Kim2018Multiple, Gao2018Blind} have shown the effectiveness of using multi-level features extracted from CNN at different depths. It inspires us to integrate the quality-related features from the attention-based spherical feature extraction module into a multi-level representation. With the multi-level representation, our approach can capture the quality degradation at different levels of granularity, benefiting BVQA on 360{\textdegree} video. Specifically, taking the features $\textbf{I}_t^{\rm LL}$, $\textbf{I}_t^{\rm ML}$ and $\textbf{I}_t^{\rm HL}$ as input, the goal of the multi-level feature re-scaling module is to re-scale the spatial size of $\textbf{I}_t^{\rm LL}$, $\textbf{I}_t^{\rm ML}$ and $\textbf{I}_t^{\rm HL}$, such that these multi-level feature maps are with the same size. Considering the trade-off between computational complexity and information integrity, we perform the down-sampling and up-sampling operations on $\textbf{I}_t^{\rm LL}$ and $\textbf{I}_t^{\rm HL}$ to keep their sizes the same as $\textbf{I}_t^{\rm ML}$. Then, a 1$\times$1 convolution layer is added behind the down-sampling and up-sampling operations to reduce the channel dimension of multi-level features. Finally, we can obtain the re-scaled multi-level features $\textbf{I}_t^{\rm LL^{\prime}}$, $\textbf{I}_t^{\rm ML^{\prime}}$ and $\textbf{I}_t^{\rm HL^{\prime}}$, and feed them into the next module of selective feature integration.

\begin{figure}[!tb]
	\begin{center}
		\includegraphics[width=1\linewidth]{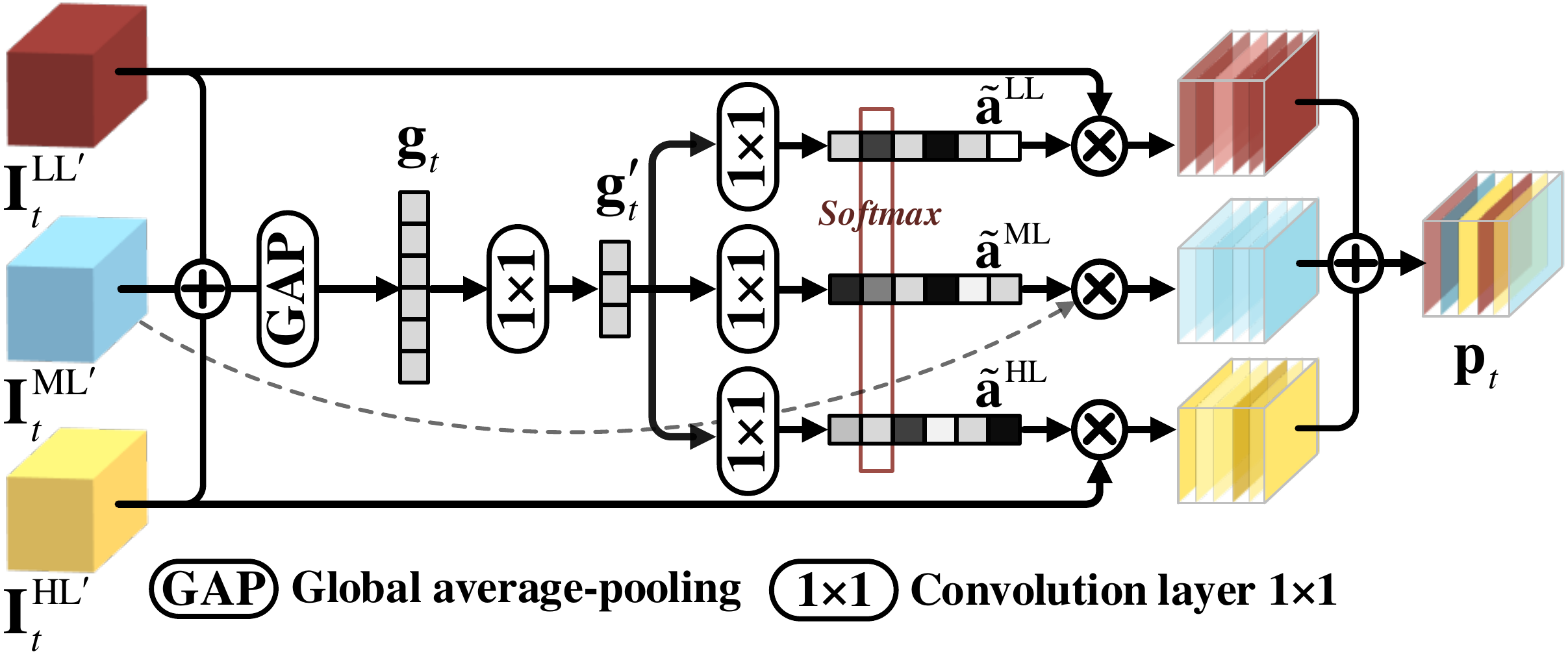}
	\end{center}
	\vspace{-0.8em}
	\caption{Structure of the proposed selective feature integration module in the SPAQ sub-net. }
	\vspace{-1.6em}
	\label{SFI}	
\end{figure}

\textbf{Selective feature integration module}. The selective feature integration module is designed for fusing the features from multiple levels, via adaptively selecting the different levels of quality degradation with the self-attention mechanism \cite{Hu2018Squeeze,li2019selective}. The structure of this module is shown in Figure \ref{SFI}. Specifically, the input features $\textbf{I}_t^{\rm LL^{\prime}}$, $\textbf{I}_t^{\rm ML^{\prime}}$ and $\textbf{I}_t^{\rm HL^{\prime}}$ are first combined as $\textbf{I}_t^{{\prime}}$ by element-wise summation:
	\vspace{-0.3em}
	\begin{equation}
	\label{sum}
	\begin{aligned}
	\textbf{I}_t^{{\prime}} = \textbf{I}_t^{\rm LL^{\prime}} + \textbf{I}_t^{\rm ML^{\prime}} + \textbf{I}_t^{\rm HL^{\prime}}. 
	\end{aligned}
	\vspace{-0.3em}
	\end{equation} 
	Then, the global average pooling (GAP) is applied across the spatial dimension of $\textbf{I}_t^{{\prime}}\in \mathbb{R}^{H\times W\times C}$, to generate the channel-wise feature $\textbf{g}_t\in \mathbb{R}^{1\times 1\times C}$. Here, the $c$-th element of $\textbf{g}_t$ is calculated by squeezing the $c$-th channel of $\textbf{I}_t^{{\prime}}$ through spatial dimensions $H\times W$:
	\vspace{-0.3em}
	\begin{equation}
	\label{gap}
	\begin{aligned}
	\textbf{g}_{t, c} = \frac{1}{H\times W} \sum_{h=1}^{H}\sum_{w=1}^{W}\textbf{I}_{t,c}^{{\prime}}(h, w), 
	\end{aligned}
	\vspace{-0.3em}
	\end{equation} 
	where $\textbf{g}_{t, c}$ and $\textbf{I}_{t,c}^{{\prime}}$ are the $c$-th element of $\textbf{g}_t$ and the $c$-th channel of $\textbf{I}_t^{{\prime}}$, respectively. Besides, $H$, $W$ and $C$ denote the height, width and the number of channels for $\textbf{I}_t^{{\prime}}$, respectively. Subsequently, a 1$\times$1 convolution is applied to reduce the number of channels of $\textbf{g}_{t}$ and generate a compact feature representation $\textbf{g}_{t}^{{\prime}}\in \mathbb{R}^{1\times 1\times \frac{C}{r}}$, where $r$ denotes the reduction ratio. Then, $\textbf{g}_{t}^{\prime}$ passes through three parallel 1$\times$1 convolution layers for channel up-scaling, and then we can obtain $\textbf{a}^{\rm LL}$, $\textbf{a}^{\rm ML}$ and $\textbf{a}^{\rm HL}\in \mathbb{R}^{1\times 1\times C}$, as the channel-attention vectors for the features $\textbf{I}_t^{\rm LL^{\prime}}$, $\textbf{I}_t^{\rm ML^{\prime}}$ and $\textbf{I}_t^{\rm HL^{\prime}}$, respectively. To facilitate the interaction among multi-level quality features, a softmax operation is conducted on $\textbf{a}^{\rm LL}$, $\textbf{a}^{\rm ML}$ and $\textbf{a}^{\rm HL}$ along with the channel dimension, such that the refined attention vectors $\widetilde{\textbf{a}}^{\rm LL}$, $\widetilde{\textbf{a}}^{\rm ML}$ and $\widetilde{\textbf{a}}^{\rm HL}$ can be generated as follows,
	\begin{eqnarray}
	\label{softmax}
	\widetilde{\textbf{a}}_c^{\rm LL} = \frac{\text{exp}({\textbf{a}_c^{\rm LL}})}{\text{exp}({\textbf{a}_c^{\rm LL}})+\text{exp}({\textbf{a}_c^{\rm ML}})+\text{exp}({\textbf{a}_c^{\rm HL}})}, \\
	\widetilde{\textbf{a}}_c^{\rm ML} = \frac{\text{exp}({\textbf{a}_c^{\rm ML}})}{\text{exp}({\textbf{a}_c^{\rm LL}})+\text{exp}({\textbf{a}_c^{\rm ML}})+\text{exp}({\textbf{a}_c^{\rm HL}})}, \\
	\widetilde{\textbf{a}}_c^{\rm HL} = \frac{\text{exp}({\textbf{a}_c^{\rm HL}})}{\text{exp}({\textbf{a}_c^{\rm LL}})+\text{exp}({\textbf{a}_c^{\rm ML}})+\text{exp}({\textbf{a}_c^{\rm HL}})},
	\end{eqnarray}
	where $\widetilde{\textbf{a}}_c^{\rm LL}$, $\widetilde{\textbf{a}}_c^{\rm ML}$ and $\widetilde{\textbf{a}}_c^{\rm HL}$ refer to the $c$-th channel of refined attention vectors after the softmax operation, and $\text{exp}(\cdot)$ refers to the exponential function. Finally, we can obtain the final feature map $\textbf{P}_t$ by element-wise product between the multi-level features and the corresponding attention vectors as
	\begin{equation}
	\label{integarte}
	\begin{aligned}
	\textbf{P}_t = \widetilde{\textbf{a}}^{\rm LL} \cdot \textbf{I}_t^{\rm LL^{\prime}} + \widetilde{\textbf{a}}^{\rm ML} \cdot \textbf{I}_t^{\rm ML^{\prime}} + \widetilde{\textbf{a}}^{\rm HL} \cdot \textbf{I}_t^{\rm HL^{\prime}}.
	\end{aligned}
	\end{equation} 

Compared with the simple concatenation among multi-level features \cite{Yan2020Blind}, the proposed module of selective feature integration can adaptively select pivotal channels from different levels of quality features and properly integrate the multi-level feature maps. This can greatly promote the prediction accuracy on BVQA for 360{\textdegree} video, as verified in the section of experimental results.  

\begin{figure}[!tb]
	\begin{center}
		\includegraphics[width=0.961\linewidth]{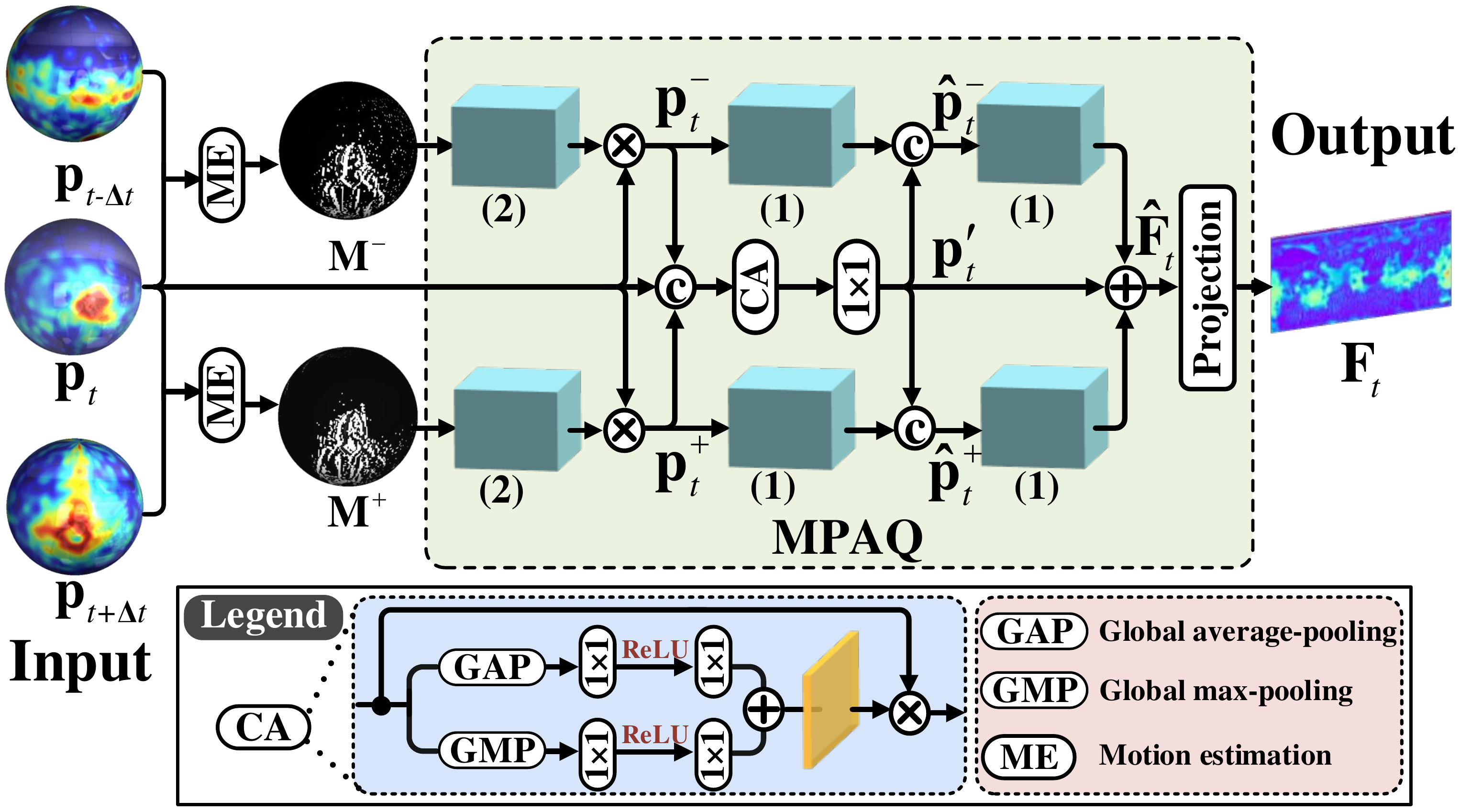}
	\end{center}
	\vspace{-0.6em}
	\caption{Architecture of the proposed ME component and MPAQ sub-net. Note that the number of output channels for all RSP blocks is 32 and the 1$\times$1 convolution is 32.}
	\vspace{-0.9em}
	\label{MPAQ}
\end{figure}

\vspace{-0.8em}
\subsection{MPAQ sub-net}

Following the SPAQ sub-net, the MPAQ sub-net is developed to leverage motion contextual information for modeling short-term quality degradation. As human perception on spatial distortion can be greatly affected by the temporal changes in videos, many works have attempted to take the motion information into consideration for the VQA task \cite{li2019viewport, xu2020viewport, manasa2016optical, seshadrinathan2007structural}. Generally speaking, these works calculate the optical flow between two adjacent frames to model motion information, and then incorporate the motion information into the spatial distortion for assessing quality. Nevertheless, the inaccurate estimation of optical flow may introduce some unexpected distortion for BVQA on 360{\textdegree} video. Besides, the computational complexity of optic flow is heavy, thus restricting the deployment  of BVQA on 360{\textdegree} video in real-world applications.

\begin{figure}[!tb]
	\begin{center}
		\includegraphics[width=0.78\linewidth]{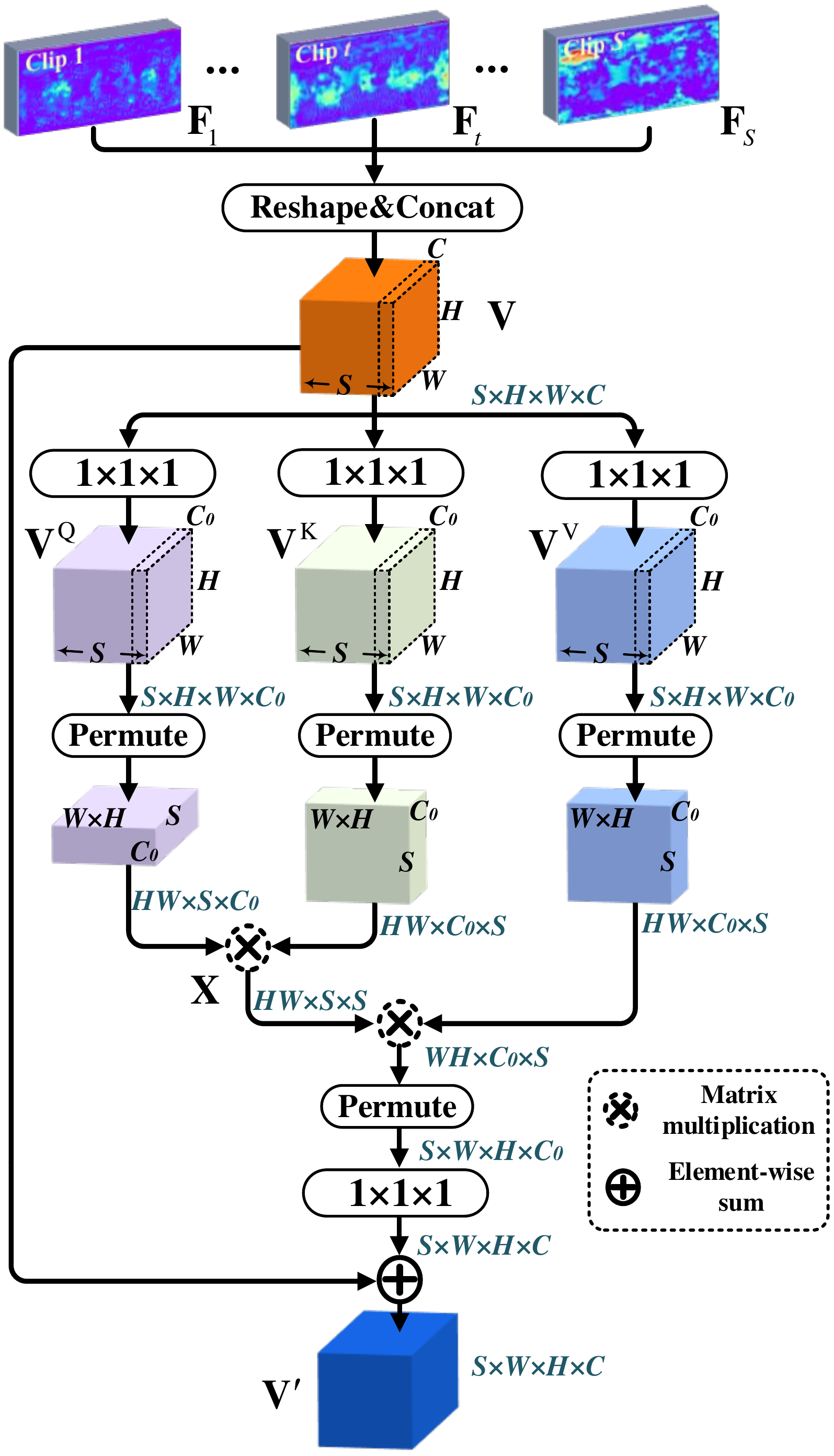}
	\end{center}
	\vspace{-1em}
	\caption{Architecture of the proposed MFTN sub-net.}
	\vspace{-1.6em}
	\label{MFTN}
\end{figure}

To this end, we propose a simple yet effective sub-net, called MPAQ, to model short-term quality degradation via incorporating motion information. The architecture of the MPAQ sub-net is illustrated in Figure \ref{MPAQ}. 
	Specifically, taking $\{\textbf{P}_{t-\Delta t}, \textbf{P}_t, \textbf{P}_{t+\Delta t}\}$ from the SPAQ sub-net as input, the motion estimation (ME) is firstly conducted through
	\begin{eqnarray}
	\label{motion}
	\begin{aligned}
	\textbf{M}^{-} = \textbf{P}_t - \textbf{P}_{t-\Delta t}, \\
	\textbf{M}^{+} = \textbf{P}_{t+\Delta t} - \textbf{P}_t,
	\end{aligned}
	\end{eqnarray}    
	where $\textbf{M}^{-}$ and $\textbf{M}^{+}$ denote the backward motion map and forward motion map, respectively. The motion maps $\textbf{M}^{-}$ and $\textbf{M}^{+}$ can capture the motion patterns from adjacent frames, complementary to the pixel-wise quality map $\textbf{P}_t$. Afterwards, $\textbf{M}^{-}$ and $\textbf{M}^{+}$, which perform as temporal masks, are flowed into a series of RSP blocks, and then multiplied with $\textbf{P}_t$ via the element-wise product as follows, 
	\begin{eqnarray}
	\label{motion1}
	\begin{aligned}
	\textbf{P}_t^{-} = \textbf{P}_t \cdot \text{RSP}_n(\textbf{M}^{-}), \\
	\textbf{P}_t^{+} = \textbf{P}_t \cdot \text{RSP}_n(\textbf{M}^{+}),
	\end{aligned}
	\end{eqnarray}
	where $\textbf{P}_t^{-}$ and $\textbf{P}_t^{+}$ denote the masked quality maps. In addition, $\text{RSP}_n(\cdot)$ refers to the operation of undergoing $n$ RSP blocks. Later, $\textbf{P}_t^{-}$, $\textbf{P}_t$ and $\textbf{P}_t^{+}$ are concatenated alongside the channel dimension, containing a large deal of information about spatial-temporal quality degradation. Then, this concatenated map passes through a channel attention and 1$\times$1 convolution layer,  obtaining the quality map $\textbf{P}_t^{\prime}$ as
	\begin{equation}
	\label{ca-11}
	\begin{aligned}
	\textbf{P}_t^{\prime} = \text{Conv}_{1\times1}\Big(\text{CA}\big(\text{Cat}(\textbf{P}_t^{-}, \textbf{P}_t, \textbf{P}_t^{+})\big)\Big), 
	\end{aligned}
	\end{equation}      
	where $\text{Cat}(\cdot)$, $\text{CA}(\cdot)$ and $\text{Conv}_{1\times1}(\cdot)$ are the operations of feature concatenation, channel attention and 1$\times$1 convolution, respectively. By using channel attention and 1$\times$1 convolution, our MPAQ sub-net can select useful information via self-attention mechanism and avoid involving too many parameters. Subsequently, $\textbf{P}_t^{-}$ and $\textbf{P}_t^{+}$ are flowed into the RSP block, and then concatenated with $\textbf{P}_t^{\prime}$ as $\hat{\textbf{P}}_t^{-}$ and $\hat{\textbf{P}}_t^{+}$:  
	\begin{eqnarray}
	\label{motion2}
	\begin{aligned}
	\hat{\textbf{P}}_t^{-} = \text{Cat}\big(\textbf{P}_t^{\prime}, \text{RSP}_n(\textbf{P}_t^{-})\big), \\
	\hat{\textbf{P}}_t^{+} = \text{Cat}\big(\textbf{P}_t^{\prime}, \text{RSP}_n(\textbf{P}_t^{+})\big).
	\end{aligned}
	\end{eqnarray}  
	Finally, the maps $\hat{\textbf{P}}_t^{-}$ and $\hat{\textbf{P}}_t^{+}$ are fed into the RSP blocks and  merged with $\textbf{P}_t^{\prime}$ by element-wise summation, to output the ultimate quality map: 
	\begin{equation}
	\label{motion_final}
	\begin{aligned}
	\hat{\textbf{F}}_t = \textbf{P}_t^{\prime} + \text{RSP}_n(\hat{\textbf{P}}_t^{-}) + \text{RSP}_n(\hat{\textbf{P}}_t^{+}).
	\end{aligned}
	\end{equation}
	Given $\hat{\textbf{F}}_t$, after the sphere-to-plane projection, we can obtain the final equi-rectangular representation of the frame-level quality map $\textbf{F}_t$.

\vspace{-0.9em}
\subsection{MFTN sub-net}

Following the MPAQ sub-net, the MFTN sub-net is designed to capture the contextual correlation among multiple inconsecutive frames, for assessing long-term quality degradation of 360{\textdegree} video. Recently, many VQA works \cite{li2019quality, li2021unified, chen2020rirnet, you2019deep} have adopted the RNN structures to integrate multi-frame features into frame-wise quality scores. In these works, average-pooling is applied on the quality scores of all frames to  obtain the final quality score. However, the frame-to-frame quality correlation is not explicitly encoded in RNN, thus failing to determine the pivotal frames that are most related with the subjective scores of 360{\textdegree} video. Moreover, since the heavy quality fluctuation exists across 360{\textdegree} video frames, it is unreasonable to utilize the average-pooling strategy for aggregating the quality score. 

In this section, we propose an MFTN sub-net to model the frame-to-frame quality correlation for 360{\textdegree} video, by designing a temporal non-local neural network on top of self-attention mechanism. The typical non-local neural network aims to learn the pairwise similarity of locations in both spatial and temporal dimensions. However, it entangles the spatial-temporal information and thus fails to capture temporal relationship across frames \cite{he2020gta}. Our MFTN sub-net is designed by disentangling the temporal attention from spatial attention, for focusing on the key frames that determine the overall video quality, which can be seen in Figure \ref{MFTN}.

Specifically, the frame-level maps $\{\textbf{F}_t\}_{t=1}^{S}$ from all sampled clips are firstly reshaped and concatenated as the video-level quality tubelet $\textbf{V} \in \mathbb{R}^{S\times H\times W\times C} $. Then, three 1$\times$1$\times$1 convolution operations with different parameters are implemented on $\textbf{V}$ for the query, key and value embedding, denoted as $\textbf{V}^{\rm Q}$, $\textbf{V}^{\rm K}$ and $\textbf{V}^{\rm V}$, respectively.
Here, the feature dimensions of $\textbf{V}^{\rm Q}$, $\textbf{V}^{\rm K}$ and $\textbf{V}^{\rm V}$ are all reduced from $C$ to $C_0$ for savig the computational complexity. Subsequently, $\textbf{V}^{\rm Q}$, $\textbf{V}^{\rm K}$ and $\textbf{V}^{\rm V}$ are reshaped and permuted to be $\textbf{V}^{\rm Q} \in \mathbb{R}^{(H\cdot W)\times S\times C_0}$, $\textbf{V}^{\rm K} \in \mathbb{R}^{(H\cdot W)\times C_0\times S}$ and $\textbf{V}^{\rm V} \in \mathbb{R}^{(H\cdot W)\times C_0\times S}$, respectively. By this means, the spatial dimension is decoupled from the temporal dimension of each embedding, and the similarity matrix $\textbf{X}$ can be obtained by
\vspace{-0.1em}
\begin{equation}
\label{s matrix}
\begin{aligned}
\textbf{X} = \textbf{V}^{\rm Q} \otimes \textbf{V}^{\rm K},
\end{aligned}
\vspace{-0.1em}
\end{equation}
where $\otimes$ denotes the matrix multiplication. Note that the similarity matrix $\textbf{X} \in \mathbb{R}^{(H\cdot W)\times S\times S}$ can be regarded as an attention map indicating frame-to-frame correlation in evaluating the overall video quality. Next, the similarity matrix $\textbf{X}$ is multiplied by the value embedding $\textbf{V}^{\rm V}$, and then undergoes a 1$\times$1$\times$1 convolution and a residual connection, as follows,
\begin{equation}
\label{s matrix1}
\begin{aligned}
\textbf{V}^{'} = \textbf{V} + \gamma\big(\rho(\textbf{V}^{\rm V} \otimes \textbf{X})\big),
\end{aligned}
\end{equation}
where $\rho(\cdot)$ and $\gamma(\cdot)$ denotes the operations of permutation and 1$\times$1$\times$1 convolution, respectively. Consequently, we can obtain the re-weighted video-level quality tubelet $\textbf{V}^{'} \in \mathbb{R}^{S\times H\times W\times C}$, which captures the long-term quality correlation across multiple inconsecutive frames.

\begin{figure}[!tb]
	\begin{center}
		\includegraphics[width=1\linewidth]{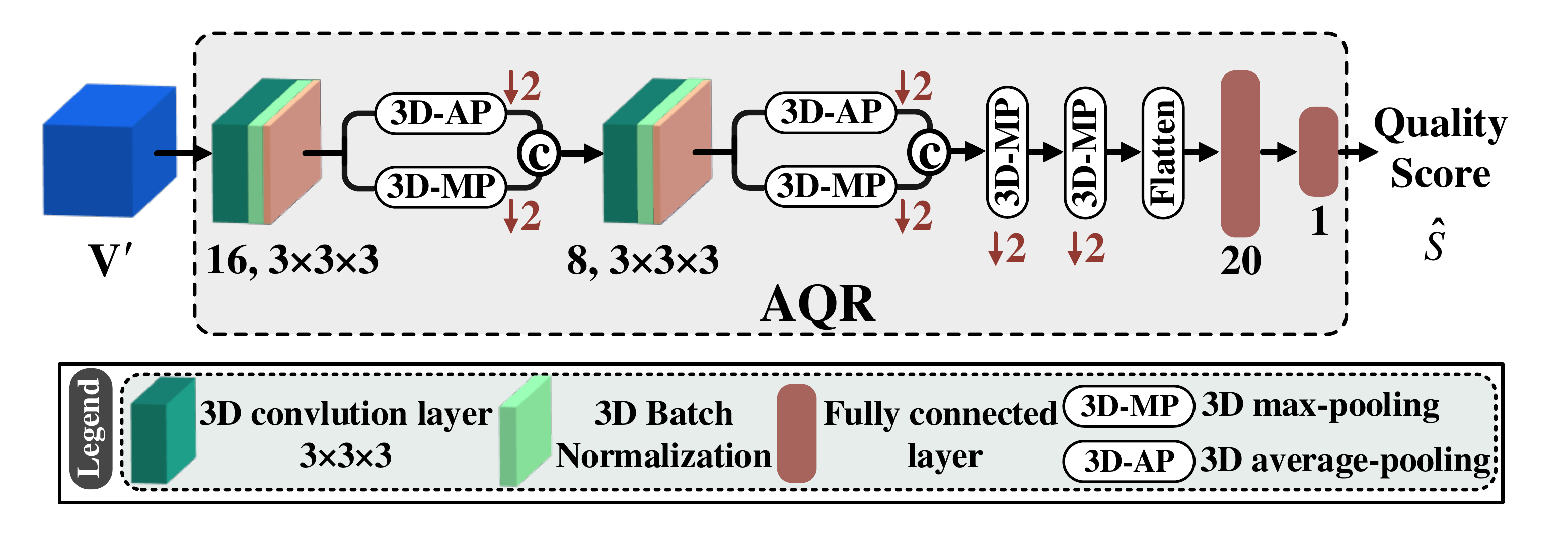}
	\end{center}
	\vspace{-0.5em}
	\caption{Architecture of our proposed AQR module. The number of output channels of two 3D convolution layers are 16 and 8, respectively. The number of output channels of two fully connected layers are 20 and 1, respectively.}
	\vspace{-1.2em}
	\label{AQR}
\end{figure}

Following the MFTN sub-net, we design the AQR module, which aims to regress the re-weighted video-level quality tubelet $\textbf{V}^{'}$ into the final quality score $\hat{s}$. The architecture of the AQR module is shown in Figure \ref{AQR}. As can be seen in this figure, the AQR module consists of two 3D convolution layers, two fully connected layers, and a series of 3D max-pooling and 3D average-pooling operations. Note that 3D max-pooling and 3D average-pooling are conducted in a complementary manner, to adaptively filter the most important features from $\textbf{V}^{'}$ for quality regression. Finally, we can obtain the predicted quality score $\hat{s}$ for the impaired 360{\textdegree} video $\textbf{I}$.

\vspace{-0.9em}	
\subsection{Training protocol.}

Now, we introduce loss function $\mathcal{L}$ for training the ProVQA model. Specifically, we utilize the mean square error (MSE) as the loss function, which measures the Euclidean distance between the vectors of predicted and ground truth quality scores, for a batch of impaired 360{\textdegree} videos. Hence, the goal is to minimize the MSE loss between the predicted quality score $\hat{s}$ and its ground truth $s$, which is formulated by
	\begin{equation}
	\label{loss_m}
	\begin{aligned}
	\mathcal L = ||\hat{s} - s||_2^2 .
	\end{aligned}
	\end{equation} 
	With the target of loss minimization, the parameters of the ProVQA model are updated using the stochastic gradient descent algorithm with the Adam optimizer \cite{kingma2014adam}.	

For training our ProVQA model, the network parameters are updated recursively with the initial learning rate of 3$\times$10$^{-4}$, and the weight decay is also applied with 5$\times$10$^{-5}$ for regularization. Moreover, we initialize the parameters of our ProVQA model by random initialization without any pre-trained weights. Our approach is trained in a total of 3$\times$10$^4$ iterations, and the batch size is set to 6 for each iteration. Considering the constraint of GPU memory, the number of sampled clips $S$ is set to 6. Other hyper-parameters are all obtained by tuning over the training set. The reduction ratio $r$ in the SPAQ sub-net is set to 8. In addition, frame interval $\Delta t$ is set to 3 for a proper motion representation.

\vspace{-0.8em}
\section{\textbf{Experiments and Results}}

In this section, we conduct extensive experiments to validate the effectiveness of the proposed ProVQA approach on BVQA for 360{\textdegree} video. First, we present the implementation details about our experiments. Then, we report the experimental results of our and other state-of-the-art approaches on BVQA for 360{\textdegree} video. Finally, the ablation studies are conducted to analyze the contribution of each component proposed in our ProVQA approach.  

\begin{table*}[!t]%
	\renewcommand\arraystretch{1.08}
	\centering%
	\caption{Comparison on BVQA performance between our and other 13 approaches, over VQA-ODV dataset. The best results are in {\color{red}{\textbf{bold}}} and the second best results are in {\color{cyan}{\underline{underline}}}.}
	\vspace{0.6em}
	\label{table:compare_1}
	\resizebox{\textwidth}{!}{%
		\normalsize
		\begin{tabular}{|c|cccc|ccccc|ccc|}%
			\hline
			\multicolumn{1}{|c|}{\multirow{2}{*}{\textbf{Approaches}}} & \multicolumn{4}{c|}{\textbf{Attributes}} & \multicolumn{5}{c|}{\textbf{Performance on VQA-ODV dataset}} & \multicolumn{3}{c|}{\textbf{Model complexity}}\\
			\cline{2-13}\multicolumn{1}{|c|}{} & No reference & Machine learning & Re-trained & Type & PLCC$\uparrow$ & SROCC$\uparrow$ & KROCC$\uparrow$ & RMSE$\downarrow$  & MAE$\downarrow$ & Running time  & Capacity & FLOPs \\
			\hline
			\multicolumn{1}{|c|}{3D-MSCN}  & \checkmark & \checkmark  & \checkmark & 2D video & 0.4680 & 0.4795 & 0.3900 & 10.4680 & 8.3551 & 17.30ms & -- & -- \\

			\multicolumn{1}{|c|}{VIDEVAL}  & \checkmark & \checkmark & \texttimes  & 2D video & 0.5288 & 0.4831 & 0.3295 & 10.0538 & 8.2828 & 85.99ms & -- & --  \\

			\multicolumn{1}{|c|}{NSTSS}  & \checkmark & \checkmark  & \checkmark & 2D video & 0.5551 & 0.5168 & 0.3581 & 9.8528 & 7.6423 & 723.27ms & -- & --  \\

			\multicolumn{1}{|c|}{VSFA}  & \checkmark & \checkmark & \checkmark & 2D video & 0.5584 & 0.4623 & 0.2997 & 9.8267 & 7.9909 & 10.00ms & 0.56M & 0.22G  \\

			\multicolumn{1}{|c|}{ST-Gabor} & \checkmark   & \checkmark & \checkmark & 2D video & 0.5713 & 0.5293 & 0.3634 & 9.7215 & 7.5426 & 705.97ms & -- & --  \\

			\multicolumn{1}{|c|}{V-MEON}  & \checkmark & \checkmark & \texttimes  & 2D video & 0.7197 & 0.7519 & 0.5427 & 8.2239 & 6.4117 & 8.31ms & 0.17M & 0.18G \\
			
			\multicolumn{1}{|c|}{TLVQM}  & \checkmark & \checkmark & \checkmark& 2D video & 0.7361 & 0.7321 & 0.5312 & 8.0175 & 6.4694 & 48.11ms & -- & -- \\
			
			\multicolumn{1}{|c|}{NR-OVQA}  & \checkmark  & \checkmark & \checkmark  & 360{\textdegree} video & 0.7598 & 0.7972 & \color{cyan}{\underline{0.6286}}  & 7.7006 & \color{cyan}{\underline{4.9496}}  &  98.52ms & 43.10M & 418.77G \\
			
			\multicolumn{1}{|c|}{S-PSNR}  & \texttimes  & \texttimes & \texttimes & 360{\textdegree} image & 0.6914 & 0.6973 & 0.4981 & 8.5582 & 6.6929 & 186.61ms & -- & -- \\
			
			\multicolumn{1}{|c|}{CPP-PSNR}  & \texttimes & \texttimes & \texttimes & 360{\textdegree} image & 0.6798 & 0.6896 & 0.4915 & 8.6875 & 6.8030 & 45.58ms & -- & -- \\
			
			\multicolumn{1}{|c|}{WS-PSNR}  & \texttimes & \texttimes & \texttimes  & 360{\textdegree} image & 0.6707 & 0.6822 & 0.4849 & 8.7854 & 6.9190 & 0.41ms & -- & -- \\

			\multicolumn{1}{|c|}{VGCN}  & \checkmark & \checkmark & \checkmark & 360{\textdegree} image & \color{cyan}{\underline{0.8032}} & \color{cyan}{\underline{0.8122}} & 0.6144 & \color{cyan}{\underline{7.0562}}  & 5.4088 & 84.67ms & 26.66M & 220.00G \\
			
			\multicolumn{1}{|c|}{MC360IQA}  & \checkmark & \checkmark & \checkmark  &  360{\textdegree} image & 0.7589 & 0.7831 & 0.6075 & 7.7134 & 5.6461 & 379.81ms & 22.40M & 22.71G \\

			\hline
			
			\multicolumn{1}{|c|}{ProVQA (Ours)} & \checkmark & \checkmark & \checkmark & 360{\textdegree} video & \textbf{\color{red}{0.9209}} & \textbf{\color{red}{0.9236}} & \textbf{\color{red}{0.7760}} & \textbf{\color{red}{4.6165}} & \textbf{\color{red}{3.1136}} & 15.32ms & 1.31M & 6.39G \\  				
			
			\hline
			
		\end{tabular}%
	}%
	\vspace{-1.2em}
\end{table*}%

\vspace{-0.8em}
\subsection{Implementation details}
\textbf{Experimental settings.}
In our experiments, the BVQA performance is evaluated over two benchmark VQA datasets of 360{\textdegree} videos, \textit{i.e.}, the VQA-ODV  \cite{xu2020viewport} and BIT360 datasets \cite{zhang2017subjective}. The details about these two datasets are introduced as follows,

\begin{itemize}
	\item \textbf{VQA-ODV dataset}. This dataset includes 540 impaired 360{\textdegree} videos from 60 reference 360{\textdegree} videos under equi-rectangular projection (ERP). The resolution of all videos covers from 4K (3840$\times$1920 pixels) to 8K (7680$\times$3840 pixels). For each reference, 9 types of impairment are applied, including three compression levels, \textit{i.e.}, QP = 27, 37 and 42, and 3 projection patterns, \textit{i.e.}, ERP, reshaped cubemap projection (RCMP) and truncated square pyramid projection (TSP). Each 360{\textdegree} video is rated by 20-30 subjects, and the subjective quality score is in a range of 0 to 100. Note that the lower score means better visual quality.  
	
	\item \textbf{BIT360 dataset}. In this dataset, there are 16 reference 360{\textdegree} videos at the resolution of 4096$\times$2048 and the frame rate of 30 frames per second (fps). The format of 360{\textdegree} reference videos is YUV420p at the bit-rate of 100 Mbps. By imposing 24 types of distortion on the reference, there are 384 impaired 360{\degree} videos produced in this dataset, the distortion of which include Gaussian noise, blur, different bitrates, \textit{etc}. Here, 23 subjects are involved to rate the subjective quality scores. The subjective quality score is in a range of 0 to 100, with the lower score referring to better visual quality. 	
\end{itemize}

In our experiments, we conduct the training process over the training set of VQA-ODV dataset, and then evaluate the performance of BVQA over the testing set of VQA-ODV dataset. In addition, we verify the generalization ability of our ProVQA approach over the BIT360 dataset. Specifically, all impaired 360{\textdegree} videos in the VQA-ODV dataset are randomly divided into training and testing sets in a ratio of 4 : 1, \textit{i.e.}, 432 training and 108 testing videos. It is worth mentioning that the subjective quality scores of VQA-ODV dataset are all normalized via being divided by 100. This normalization procedure can ease the training process and accelerate the convergence of loss. The normalized quality scores are viewed as the ground truth quality scores in the range of $\lbrack$0, 1$\rbrack$, where the higher score means worse visual quality. Furthermore, our approach is implemented based on the PyTorch framework \cite{paszke2019pytorch}, and run on two NVIDIA Tesla V100 GPUs with 32G memory. 	

\textbf{Evaluation metrics.}
The performance of BVQA on 360{\textdegree} video can be evaluated by measuring the consistency between the predicted quality score by ProVQA and its ground truth quality score. To this end, we adopt five standard evaluation metrics, including Pearson linear correlation coefficient (PLCC), Spearman rank order correlation coefficient (SROCC), Kendall rank-order correlation coefficient (KROCC), root mean squared error (RMSE) and mean absolute error (MAE). In particular, SROCC and KROCC measure the prediction monotonicity, while PLCC, RMSE and MAE measure the prediction accuracy. For PLCC, SROCC and KROCC, the larger values imply higher correlation, while the smaller values of RMSE and MAE mean fewer prediction errors. Before calculating the above metrics, we employ a logistic function to fit the predicted quality scores to their corresponding ground truth, such that the fitted scores of all approaches have the same scale as their ground truth.


\begin{figure*}[!tb]
	\begin{center}
		\includegraphics[width=0.87\linewidth]{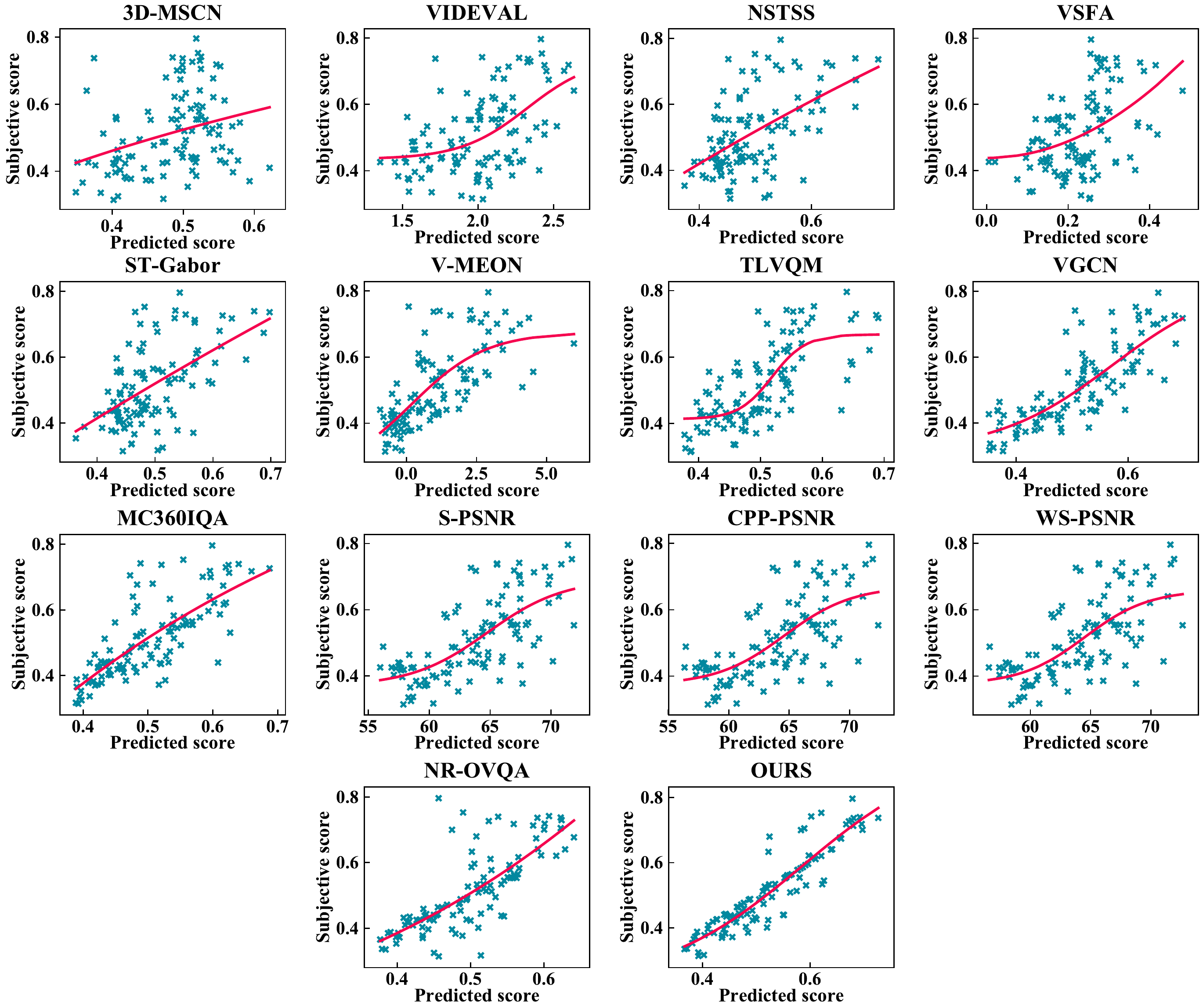}
	\end{center}
	\vspace{-1.em}
	\caption{Scatter plots for the pairs of subjective and predicted quality scores of 13 compared approaches and ours, over the test set of ODV-VQA dataset.}
	\label{subject_exp}
	\vspace{-1.4em}
\end{figure*}

\vspace{-1.2em}
\subsection{Performance evaluation}

\textbf{Compared approaches.}
Here, we compare the performance between our ProVQA approach and other 8 latest state-of-the-art approaches for BVQA on 360{\textdegree} video or 2D video. The compared approaches include 3D-MSCN \cite{dendi2020no}, VIDEVAL \cite{tu2021ugc}, NSTSS \cite{dendi2020no}, VSFA \cite{li2019quality}, ST-Gabor \cite{dendi2020no}, V-MEON \cite{liu2018end}, TLVQM \cite{korhonen2019two} and NR-OVQA \cite{chai2021blind}. Among them, NR-OVQA{\footnote{The code of NR-OVQA is obtained by asking from the author (747866472@qq.com).}} is the only existing public BVQA approach on 360{\textdegree} video, while other approaches are devoted to BVQA on 2D video. Table \ref{table:compare_1} shows the details about the attributes of these approaches. Note that all the compared approaches are based on machine learning paradigm, in which VSFA, V-MEON and NR-OVQA are deep learning-based approaches. For fair comparison, all compared approaches except VIDEVAL and V-MEON are re-trained using the same training set of VQA-ODV dataset as ours, since only the testing codes of VIDEVAL and V-MEON are available in public. Then, we evaluate the BVQA performance using the trained models over the same testing set.

Since there exists only one BVQA approach for 360{\textdegree} video so far, we further compare five 360{\textdegree} image VQA approaches,  including VGCN \cite{xu2020blind}, MC360IQA \cite{sun2019mc360iqa}, S-PSNR \cite{Yu2015A}, CPP-PSNR \cite{Zakharchenko2016Video} and WS-PSNR \cite{Sun2017Weighted}. Among them, VGCN and MC360IQA are two latest CNN-based BVQA approaches for 360{\textdegree} image, while S-PSNR, CPP-PSNR and WS-PSNR are PSNR-related FR VQA approaches evaluating the pixel-level fidelity of 360{\textdegree} image. Note that all of these approaches take the sphere perception characteristics into account, in assessing the quality degradation for 360{\textdegree} image. Likewise, we re-train the networks of VGCN and MC360IQA on the same training set as ours, and then evaluate their BVQA performance. Since these approaches apply for 360{\textdegree} image not video, we make each frame of 360{\textdegree} video as a training sample during the training stage. During the test stage, the final predicted quality score of each 360{\degree} video is obtained via averaging the quality scores of all frames.  

\textbf{Quantitative results.}
Table \ref{table:compare_1} tabulates the performance of BVQA for our ProVQA approach and 13 other approaches, in terms of PLCC, SROCC, KROCC, RMSE and MAE. The results in this table are obtained over all testing 360{\textdegree} videos in the VQA-ODV dataset. From this table, we can see that our approach performs significantly better than all other approaches in terms of 5 metrics. To be more specific, our approach achieves at least 0.118, 0.111, 0.147, 2.440 and 1.836 gains in PLCC, SROCC, KROCC, RMSE and MAE, respectively. In particular, the proportion of improvement over the second-best approach VGCN or NR-OVQA is 14.69\% in PLCC, 13.72\% in SROCC, 23.45\% in KROCC, 34.58\% in RMSE, and 37.09\% in MAE. In summary, the quantitative results validate the effectiveness of our ProVQA approach for BVQA on 360{\textdegree} video. 

\textbf{Scatter plots.}
Figure \ref{subject_exp} plots the scatters of the predicted quality scores versus their ground truth scores over all 108 impaired 360{\degree} videos in the test set, to visualize their correlation. Additionally, the logistic fitting curves regressed from the predicted scores are also shown in this figure. Generally speaking, the intensive scatter points close to the fitting curves are of little error, indicating the high correlation between the predicted and ground truth quality scores. It can be obviously observed from Figure \ref{subject_exp} that the predicted scores by our ProVQA approach have much higher correlation with the ground truth quality scores, compared with all other approaches. Moreover, this figure also shows that our approach can maintain excellent prediction monotonicity and accuracy for BVQA on 360{\textdegree} video. Therefore, we can conclude that our ProVQA approach is able to accurately predict the subjective quality scores of 360{\textdegree} video, performing considerably better than other approaches. 

\textbf{Model complexity.}
We further compare our and other comparison approaches in terms of the running time, number of parameters and floating point operations (FLOPs). For the running time, we execute all experiments over the test set of the ODV-VQA dataset with the same hardware configuration. Table \ref{table:compare_1} reports the  average running time for each frame of test 360{\textdegree} video. We can observe that our ProVQA approach consumes less  execution time than most comparison approaches. Since VSFA, V-MEON, NR-OVQA, VGCN and MC360IQA are CNN-based approaches, we further measure their FLOPs and parameter numbers. Tables \ref{table:compare_1} shows that our approach requires around 1.31M parameters, which is close to those of VSFA and V-MEON, but our approach achieves significantly better performance than these two approaches. Compared with NR-OVQA (43.10M), VGCN (26.66M) and MC360IQA (22.40M), our approach has dramatically less parameter number, but achieves better BVQA performance.
Similar results can be found for FLOPs. All above results validate the high efficiency of our ProVQA approach for BVQA on 360{\textdegree} video.

\begin{figure*}[!tb]
	\begin{center}
		\includegraphics[width=0.87\linewidth]{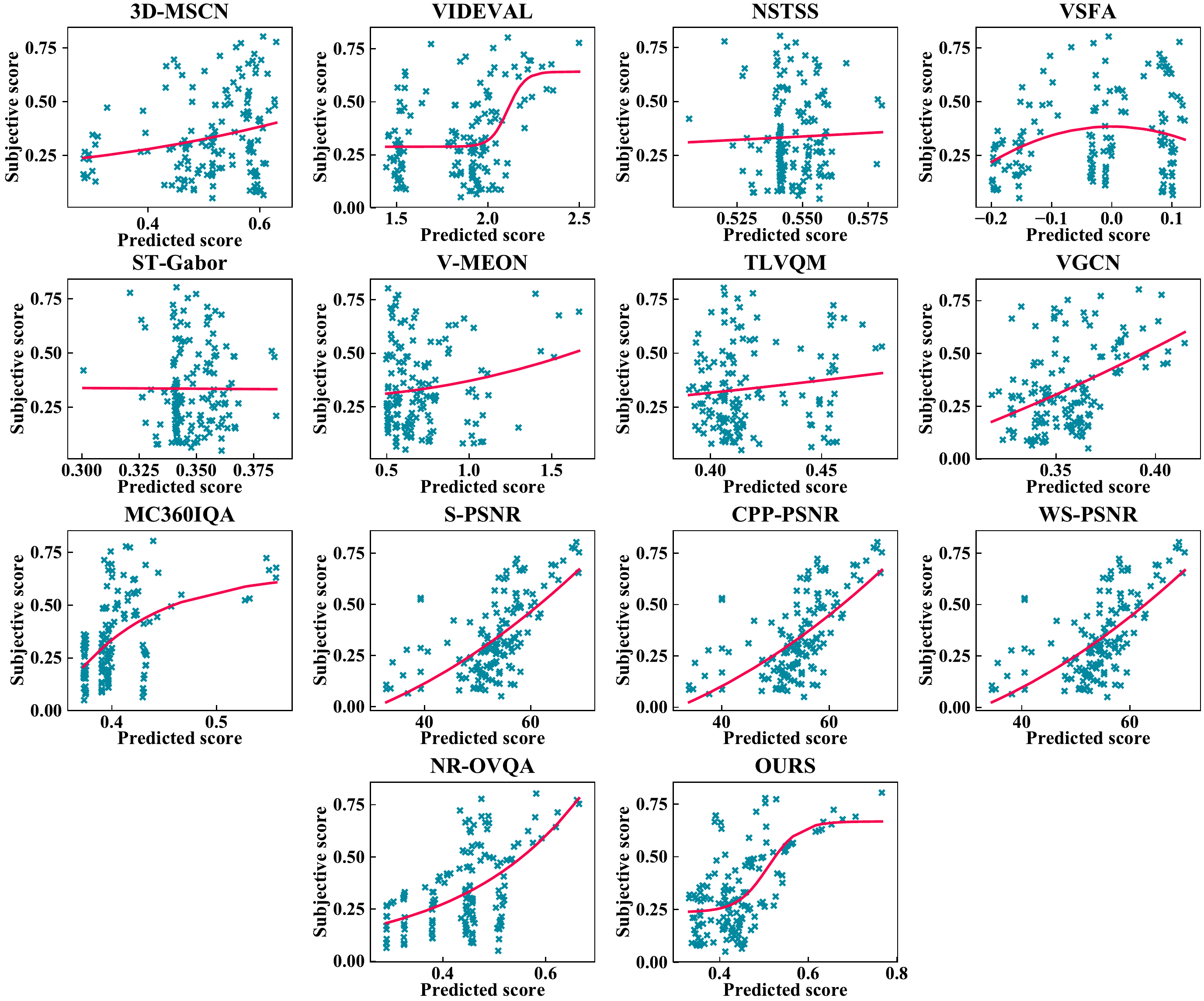}
	\end{center}
	\vspace{-1.em}
	\caption{Scatter plots for the pairs of subjective and predicted quality scores of 13 compared approaches and ours, over the test set of BIT360 dataset.}
	\label{subject_exp_2}
	\vspace{-1.5em}
\end{figure*}

\textbf{Generalization analysis.}
	We further evaluate the generalization ability of our ProVQA approach for BVQA on 360{\textdegree} video. To this end, we compare the performance of our approach and 13 other approaches over another dataset BIT360. Here, we select 144 impaired 360{\textdegree} videos as the test set, where all 24 distortion types in the dataset are involved. Note that our and all other learning-based approaches, trained over the VQA-ODV dataset, are directly applied to assess the quality of 144 impaired 360{\textdegree} videos in the BIT360 dataset, without any re-training procedure.  Table \ref{table:BIT360} reports the results of PLCC, SROCC, KROCC, RMSE and MAE over the BIT360 dataset. As shown in this table, our approach again outperforms all other approaches for the metrics of PLCC, RMSE and MAE, with at least 0.064, 0.947 and 1.059 improvement. For KROCC and SROCC, our approach is slightly inferior to the FR VQA approaches of S-PSNR and CPP-PSNR, which make use of the reference 360{\textdegree} videos during the test stage. Furthermore, we also plot the scatters of predicted and ground truth quality scores, as well as their corresponding fitting curves in Figure \ref{subject_exp_2}. The figure reveals that our approach achieves more accurate results in predicting the ground truth of subjective quality scores, compared with other approaches.    
	Generally speaking, the results on the BIT360 dataset indicate the high generalization ability of our ProVQA approach for the task of BVQA on 360{\textdegree} video.

\begin{table}[!tb]
	\renewcommand\arraystretch{1.13}
	\centering
	\caption{Generalization performance of our and other 13 approaches over BIT360 dataset. The best results are in {\color{red}{\textbf{bold}}} and the second best results are in {\color{cyan}{\underline{underline}}}.}
	\vspace{.8em}
	\resizebox{1\linewidth}{!}{
		\footnotesize
		\begin{tabular}{|c|ccccc|}
			\hline
			\multicolumn{1}{|c|}{\multirow{2}{*}{\textbf{Approaches}}}  & \multicolumn{5}{c|}{\textbf{BVQA performance on BIT360 dataset}}  \\
			\cline{2-6}\multicolumn{1}{|c|}{} & PLCC$\uparrow$ & SROCC$\uparrow$ & KROCC$\uparrow$ & RMSE$\downarrow$  & MAE$\downarrow$\\
			\hline
			\multicolumn{1}{|c|}{3D-MSCN}    & 0.2244 & 0.1628 & 0.1206 & 18.7066 & 15.1619 \\			
			\multicolumn{1}{|c|}{VIDEVAL}    & 0.5143 & 0.2943 & 0.1962 & 16.4634 & 13.0862 \\			
			\multicolumn{1}{|c|}{NSTSS}   & 0.1682 & 0.1439 & 0.1177 & 18.9230 & 15.7346 \\			
			\multicolumn{1}{|c|}{VSFA}   & 0.3281 & 0.3151 & 0.2551 & 18.1336 & 15.0755 
			\\		
			\multicolumn{1}{|c|}{ST-Gabor}   & 0.2136 & 0.1819 & 0.1490 & 18.7534 & 15.5493  \\		
			\multicolumn{1}{|c|}{V-MEON} & 0.2888 & 0.2514 & 0.2060 & 18.3784 & 15.1204 
			\\		
			\multicolumn{1}{|c|}{TLVQM}   & 0.2308 & 0.0780 & 0.0446 & 18.6783 & 15.1515 
			\\
			\multicolumn{1}{|c|}{NR-OVQA}  & 0.5751 & 0.4838 & 0.3502 & 15.7038 & 12.6590 
			\\
			\multicolumn{1}{|c|}{S-PSNR}   & \color{cyan}{\underline{0.5756}}  &\textbf{\color{red} 0.5883} & \textbf{\color{red} 0.4820} & \color{cyan}{\underline{15.6979}}   & \color{cyan}{\underline{12.6392}}   
			\\	
			\multicolumn{1}{|c|}{CPP-PSNR}  & 0.5675 &\color{cyan}{\underline{0.5810}}  & \color{cyan}{\underline{0.4760}} & 15.8056 & 12.7955 
			\\
			\multicolumn{1}{|c|}{WS-PSNR} & 0.5657 & 0.5776 & 0.4732 & 15.8296 & 12.8337 
			\\
			\multicolumn{1}{|c|}{VGCN}   & 0.5320 & 0.4221 & 0.3179 & 16.2545 & 13.3048  \\
			\multicolumn{1}{|c|}{MC360IQA}  & 0.5139 & 0.4605 & 0.3360 & 16.4681 & 12.7982  \\
			\hline
			\multicolumn{1}{|c|}{ProVQA (Ours)} & \textbf{\color{red}{0.6399}} & 0.5038 & 0.3605 & \textbf{\color{red}{14.7513}} & \textbf{\color{red}{11.5803}} \\  				
			\hline
		\end{tabular}
	}
	\label{table:BIT360}
	\vspace{-1.8em}
\end{table}

\begin{table*}[!t]%
	\renewcommand\arraystretch{1.32}
	\centering%
	\caption{Ablation on the backbone and operations of the SPAQ sub-net, over the VQA-ODV and BIT360 datasets. The best results are in {\color{red}{\textbf{bold}}}, the second and third best are in {\color{cyan}{\underline{underline}}} and {\color{green}{\underline{underline}}}, respectively.}
	\vspace{0.6em}
	\label{table:abl_1}
	
	\resizebox{\textwidth}{!}{%
		\footnotesize
		\begin{tabular}{|c|cc|ccccc|ccccc|}%
			\hline
			\multicolumn{1}{|c|}{\multirow{2}{*}{\textbf{Backbones}}} & \multicolumn{2}{c|}{\textbf{Operations}} & \multicolumn{5}{c|}{\textbf{Ablation on the VQA-ODV dataset}} & \multicolumn{5}{c|}{\textbf{Ablation on the BIT360 dataset}}\\
			\cline{2-13}\multicolumn{1}{|c|}{} & Long skip & \multicolumn{1}{c|}{Spatial attention} & PLCC $\uparrow$ & SROCC $\uparrow$ & KROCC $\uparrow$ & RMSE $\downarrow$  & \multicolumn{1}{c|}{MAE $\downarrow$} & PLCC $\uparrow$ & SROCC $\uparrow$ & KROCC $\uparrow$ & RMSE $\downarrow$ & \multicolumn{1}{c|}{MAE $\downarrow$}\\
			\hline			
			\multirow{4}*{SphereNet} & \xmark  & \xmark  
			& 0.4209 & 0.4858 & 0.3652 & 10.7447 & 8.6616 & 0.4835 & 0.3644 & 0.2469 & 16.8032 & 12.9021\\
			& \xmark  & \cmark 
			& \color{cyan}{\underline{0.8328}} & \color{cyan}{\underline{0.8286}} & \color{cyan}{\underline{0.6594}} & \color{cyan}{\underline{6.5563}}  & \color{cyan}{\underline{4.3767}}  & 0.5397 & 0.4190 & 0.2976 & 16.1600 & 13.0177\\
			& \cmark  & \xmark 
			& 0.7279 & 0.7822 & 0.5985 & 8.1224 & 6.1742 & \color{cyan}{\underline{0.5674}} & \color{cyan}{\underline{0.4714}} & \textbf{\color{red}{0.3847}} & \color{cyan}{\underline{15.8066}} & \color{green}{\underline{12.1721}}\\
			& \cmark  & \cmark 	 
			& \textbf{\color{red}{0.9209}} & \textbf{\color{red}{0.9236}} & \textbf{\color{red}{0.7760}} & \textbf{\color{red}{4.6165}} & \textbf{\color{red}{3.1136}} & \textbf{\color{red}{0.6399}} & \textbf{\color{red}{0.5038}} & \color{cyan}{\underline{0.3605}} & \textbf{\color{red}{14.7513}} & \textbf{\color{red}{11.5803}}\\  									
			\hline			
			\multirow{4}*{ResNet} & \xmark  & \xmark  
			& 0.4605 & 0.3163 & 0.2174 & 10.5147 & 8.6497 & \color{green}{\underline{0.5661}} & \color{green}{\underline{0.4379}}  & 0.3223 & \color{green}{\underline{15.8246}} & \color{cyan}{\underline{12.1485}}\\
			& \xmark  & \cmark 
			& 0.7164 & 0.6998 & 0.5244 & 8.2644 & 6.2395 & 0.5286 & 0.2250 & 0.1654 & 16.2957 & 12.6806\\
			& \cmark  & \xmark 
			& 0.7166 & 0.6702 & 0.5199 & 8.2619 & 6.5054 & 0.4381 & 0.2805 & 0.2206 & 17.2565 & 14.0969\\
			& \cmark  & \cmark 	
			& \color{green}{\underline{0.8014}} & \color{green}{\underline{0.8000}} & \color{green}{\underline{0.6203}} & \color{green}{\underline{7.0843}} & \color{green}{\underline{5.0916}} & 0.4681 & 0.4295 & \color{green}{\underline{0.3484}} & 16.9653 & 13.6800 \\ 			
			\hline
			
		\end{tabular}%
	}%
	\vspace{-1.6em}
\end{table*}%

\begin{table}[!tb]
	\renewcommand\arraystretch{1.2}
	\centering
	\caption{Ablation on the selective feature integration module, over the VQA-ODA and BIT360 datasets.}
	\vspace{.8em}
	\resizebox{1\linewidth}{!}{
		\Large
		\begin{tabular}{|c|c|cccc|}
			\hline
			\multirow{1}{*}{\textbf{Metrics}} & \multicolumn{5}{c|}{\textbf{Ablation on the VQA-ODV dataset}} \\
			\hline
			& \multicolumn{1}{c|}{Ours} & Sum (w/o CA)  & Sum (w/ CA) & Concat (w/o CA) &  Concat (w/ CA)  \\ 
			\cline{2-6}
			\multirow{1}{*}{PLCC $\uparrow$} & \textbf{\color{red}{0.9209}} & \color{green}{\underline{0.8755}} & 0.8316  & 0.8400  & \color{cyan}{\underline{0.8983}}  \\
			\multirow{1}{*}{SROCC $\uparrow$} & \multicolumn{1}{c|}{\textbf{\color{red}{0.9236}}} & \color{green}{\underline{0.8757}}  & 0.8268 & 0.8516 & \color{cyan}{\underline{0.8929}}    \\
			\multirow{1}{*}{KROCC $\uparrow$} & \multicolumn{1}{c|}{\textbf{\color{red}{0.7760}}}  & 
			\color{green}{\underline{0.7196}}  & 0.6407  & 0.6881  & \color{cyan}{\underline{0.7428}}    \\
			\multirow{1}{*}{RMSE $\downarrow$} & \multicolumn{1}{c|}{\textbf{\color{red}{4.6165}}} & \color{green}{\underline{5.7229}}   & 6.5793  & 6.4264 & \color{cyan}{\underline{5.2034}}  \\
			\multirow{1}{*}{MAE $\downarrow$} & \multicolumn{1}{c|}{\textbf{\color{red}{3.1136}}} & 
			\color{green}{\underline{3.7578}}  & 4.7680 & 4.3148 & \color{cyan}{\underline{3.6128}}    \\
			\hline
			\multirow{1}{*}{\textbf{Metrics}} & \multicolumn{5}{c|}{\textbf{Ablation on the BIT360 dataset}} \\
			\hline
			& \multicolumn{1}{c|}{ProVQA} & Sum (w/o CA)    & Sum (w/ CA) & Concat (w/o CA) &  Concat (w/ CA)  \\ 
			\cline{2-6}
			\multirow{1}{*}{PLCC $\uparrow$} & \multicolumn{1}{c|}{\textbf{\color{red}{0.6399}}} & 0.3847  & \color{cyan}{\underline{0.6288}}   & 0.5131  & \color{green}{\underline{0.5407}}   \\
			\multirow{1}{*}{SROCC $\uparrow$} & \multicolumn{1}{c|}{\textbf{\color{red}{0.5038}}} & 0.3221 & 0.3913  & \color{green}{\underline{0.4544}}   & \color{cyan}{\underline{0.4620}}   \\
			\multirow{1}{*}{KROCC $\uparrow$} & \multicolumn{1}{c|}{\textbf{\color{red}{0.3605}}} & 0.2253  & 0.2791 & \color{green}{\underline{0.3267}} & \color{cyan}{\underline{0.3294}} \\
			\multirow{1}{*}{RMSE $\downarrow$} & \multicolumn{1}{c|}{\textbf{\color{red}{14.7513}}}  & 17.7187  & \color{cyan}{\underline{14.9262}}  & 16.4773  & \color{green}{\underline{16.1478}} \\
			\multirow{1}{*}{MAE $\downarrow$} & \multicolumn{1}{c|}{\textbf{\color{red}{11.5803}}} & 14.7944  & \color{cyan}{\underline{11.9131}} & 13.5795 & \color{green}{\underline{13.0072}} \\
			\hline	
		\end{tabular}
	}
	\label{table:abl_2}
	\vspace{-1.6em}
\end{table} 

\vspace{-0.7em}
\subsection{\textbf{Ablation analysis}}

In the following, we conduct a series of ablation experiments with different network settings over both ODV-VQA and BIT360 datasets. Furthermore, we analyze the influence of hyper-parameters on the performance for 360{\textdegree} video.

\textbf{Ablation on the SPAQ sub-net.}
In the SPAQ sub-net, the SphereNet backbone is a key component for modeling the spatial degradation of  each 360{\textdegree} frame. Additionally, the operations of long skip and spherical spatial attention are designed to improve the BVQA performance. We therefore investigate the effectiveness of the backbone and the operations of long skip and spherical spatial attention. To this end, we first substitute the SphereNet backbone in the SPAQ sub-net with the traditional ResNet backbone. Concretely, all convolution layers of SphereNet in the SPAQ sub-net is replaced by standard convolution layers in the residual block of ResNet. Under different backbones, we further analyze the impact with or without the operations of long skip and spatial attention in the SPAQ sub-net. Consequently, there are 8 ablation settings as can be seen in Table \ref{table:abl_1}. Table \ref{table:abl_1} further reports the performance of our ProVQA approach with these 8 ablation settings, which is trained and evaluated over the datasets of VQA and BIT360. We can observe that the performance significantly degrades, when our SphereNet backbone is substituted by the ResNet backbone, regardless of whether the long skip and spatial attention are used. In another side, the adoption of long skip and spatial attention in both backbones have a large gain over the settings of removing them. In summary, the ablation experiments show that our SphereNet backbone, long skip and spatial attention operations all have positive impacts on BVQA for 360{\textdegree} video. 

Furthermore, in the SPAQ sub-net, the proposed module of selective feature integration is effective in integrating multi-level features for BVQA. Hence, we conduct the ablation experiments to validate its effectiveness on BVQA for 360{\textdegree} video. In our experiments, we replace the selective feature integration module with four different modules that employ naive mechanisms for multi-level feature integration. Specifically, the ablation implementation includes multi-level feature summation and concatenation with or without channel attention, denoted as Sum (w/ CA), Sum (w/o CA), Concat (w/ CA) and Concat (w/o CA). Table \ref{table:abl_2} shows the performance of our ablation over both ODV-VQA and BIT360 datasets. We can see that our selective feature integration module outperforms all other modules over two datasets. To be specific, this module can improve PLCC, SROCC and KROCC by 0.023, 0.031 and 0.033 against the Concat (w/ CA) over the VQA-ODV dataset. Similar improvement can be achieved against sum (w/o CA). This indicates the effectiveness of our selective feature integration module on BVQA for 360{\textdegree} video.  

\begin{figure}[!tb]
	\begin{center}
		\includegraphics[width=0.93\linewidth]{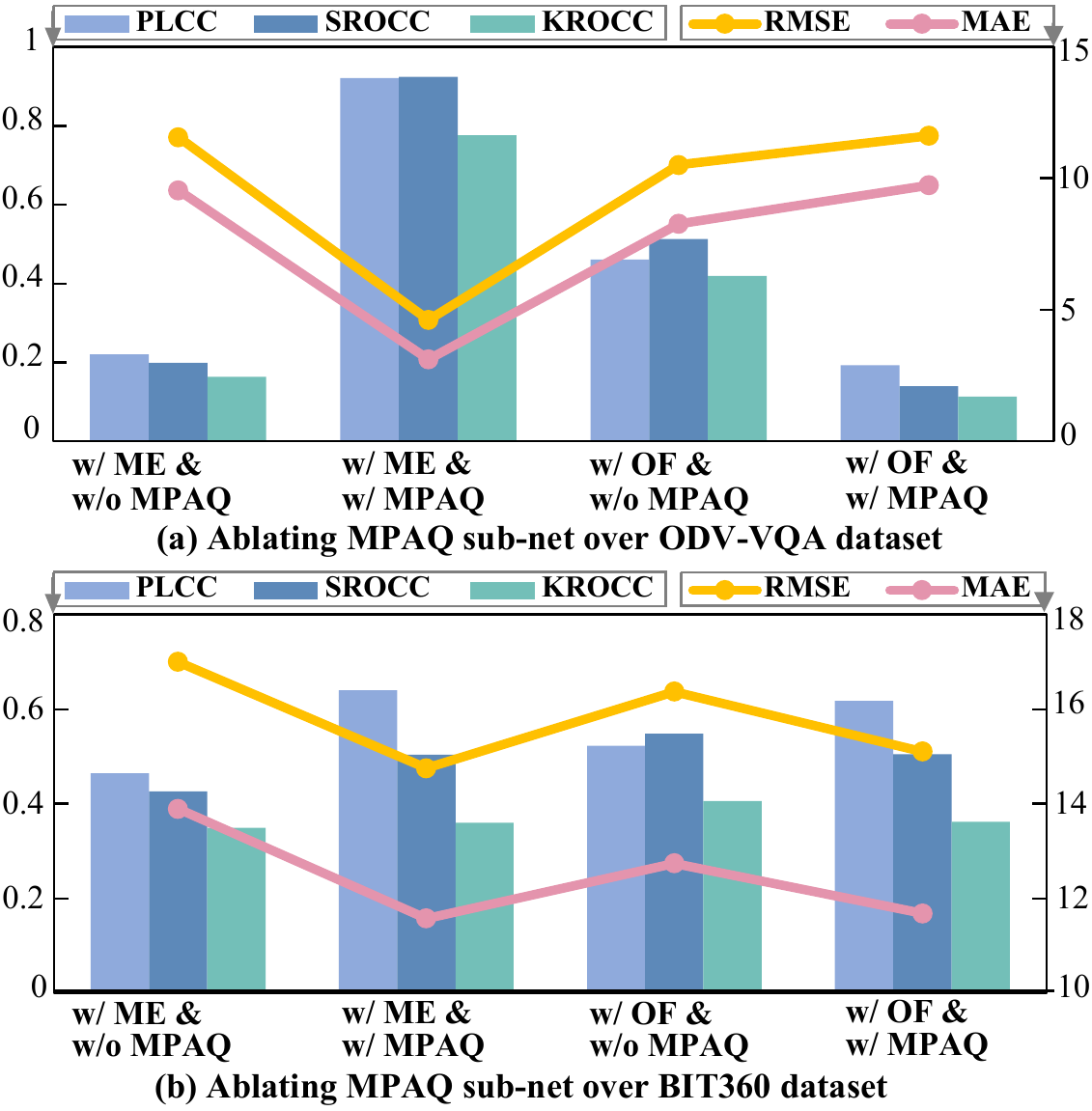}
	\end{center}
	\vspace{-1.5em}
	\caption{Ablation on ME component and MPAQ sub-net.}
	\label{abl_MPAQ}
	\vspace{-1.6em}
\end{figure}

\textbf{Ablation on the MPAQ sub-net.}
In our ProVQA approach, the MPAQ sub-net is developed to incorporate motion contextual information from the ME component for modeling short-term quality degradation. We first evaluate the effectiveness of the ME component by replacing its output motion maps by the optical flow maps.
Specifically, we replace motion maps $\textbf{M}^{-}$ and $\textbf{M}^{+}$ with two optical flow maps and input them into the MPAQ sub-net, denoted as w/ OF \& w/ MPAQ in Figure \ref{abl_MPAQ}. From this figure, we can observe that the performance is dramatically declined by 79\%, 85\% and  85\% in terms of PLCC, SROCC and KROCC, when utilizing the optical flow maps. Furthermore, the effectiveness of the MPAQ sub-net is explored by removing it from our ProVQA approach and its optical flow version, called w/ ME \& w/o MPAQ and w/ OF \& w/o MPAQ, respectively. We can see that the removal of the MPAQ sub-net results in the severe degradation of the BVQA performance for our ProVQA approach. All above results validate the effectiveness of our motion estimation component and MPAQ sub-net.      

\begin{table}[!tb]
	\renewcommand\arraystretch{1.26}
	\centering
	\caption{Ablation on the MFTN module, over the VQA-ODA and BIT360 datasets.}
	\vspace{.8em}
	\resizebox{1\linewidth}{!}{
		\tiny
		\begin{tabular}{|c|ccccc|}
			\hline
			\multirow{1}{*}{\textbf{Metrics}} & \multicolumn{5}{c|}{\textbf{Ablation on VQA-ODV dataset}} \\
			\hline
			& \multicolumn{1}{c|}{MFTN} & ConvLSTM  & LSTM & RNN & C3D  \\ 
			\cline{2-6}
			\multirow{1}{*}{PLCC $\uparrow$} & \multicolumn{1}{c|}{\textbf{\color{red}{0.9209}}} & \color{green}{\underline{0.5343}} & 0.3578 & 0.3496  & \color{cyan}{\underline{0.9072}}   \\
			\multirow{1}{*}{SROCC $\uparrow$} & \multicolumn{1}{c|}{\textbf{\color{red}{0.9236}}} & \color{green}{\underline{0.5119}}  & 0.3025 & 0.3018 & \color{cyan}{\underline{0.9018}}    \\
			\multirow{1}{*}{KROCC $\uparrow$} & \multicolumn{1}{c|}{\textbf{\color{red}{0.7760}}}  & \color{green}{\underline{0.3662}}  & 0.2106  & 0.2094  & \color{cyan}{\underline{0.7484}} \\
			\multirow{1}{*}{RMSE $\downarrow$} & \multicolumn{1}{c|}{\textbf{\color{red}{4.6165}}} & \color{green}{\underline{10.0128}}  & 11.8451  & 11.8892 &  \color{cyan}{\underline{4.9841}}   \\
			\multirow{1}{*}{MAE $\downarrow$} & \multicolumn{1}{c|}{\textbf{\color{red}{3.1136}}} & \color{green}{\underline{7.6434}}  & 9.9230 & 9.9782 & \color{cyan}{\underline{3.7254}}    \\
			\hline
			\multirow{1}{*}{\textbf{Metrics}} & \multicolumn{5}{c|}{\textbf{Ablation on BIT360 dataset}} \\
			\hline
			& \multicolumn{1}{c|}{Ours} & ConvLSTM  & LSTM & RNN & C3D  \\ 
			\cline{2-6}
			\multirow{1}{*}{PLCC $\uparrow$} & \multicolumn{1}{c|}{\textbf{\color{red}{0.6399}}} & \color{cyan}{\underline{0.5829}}   & 0.3426 & 0.3257  & \color{green}{\underline{0.5540}}  \\
			\multirow{1}{*}{SROCC $\uparrow$} & \multicolumn{1}{c|}{\textbf{\color{red}{0.5038}}} & \color{green}{\underline{0.3577}} & 0.2753  & 0.2438 & \color{cyan}{\underline{0.4965}}  \\
			\multirow{1}{*}{KROCC $\uparrow$} & \multicolumn{1}{c|}{\textbf{\color{red}{0.3605}}} & \color{green}{\underline{0.2523}} & 0.2136 & 0.1769 & \color{cyan}{\underline{0.3545}}  \\
			\multirow{1}{*}{RMSE $\downarrow$} & \multicolumn{1}{c|}{\textbf{\color{red}{14.7513}}}  & \color{cyan}{\underline{15.5980}}  & 17.6548  & 18.5375  & \color{green}{\underline{15.9815}}  \\
			\multirow{1}{*}{MAE $\downarrow$} & \multicolumn{1}{c|}{\textbf{\color{red}{11.5803}}} & \color{cyan}{\underline{11.8877}}  & 14.1624 & 14.6239 & \color{green}{\underline{12.4928}}  \\
			\hline	
		\end{tabular}
	}
	\label{table:abl_5}
	\vspace{-0.4em}
\end{table}

\begin{figure}[!tb]
	\begin{center}
		\includegraphics[width=0.99\linewidth]{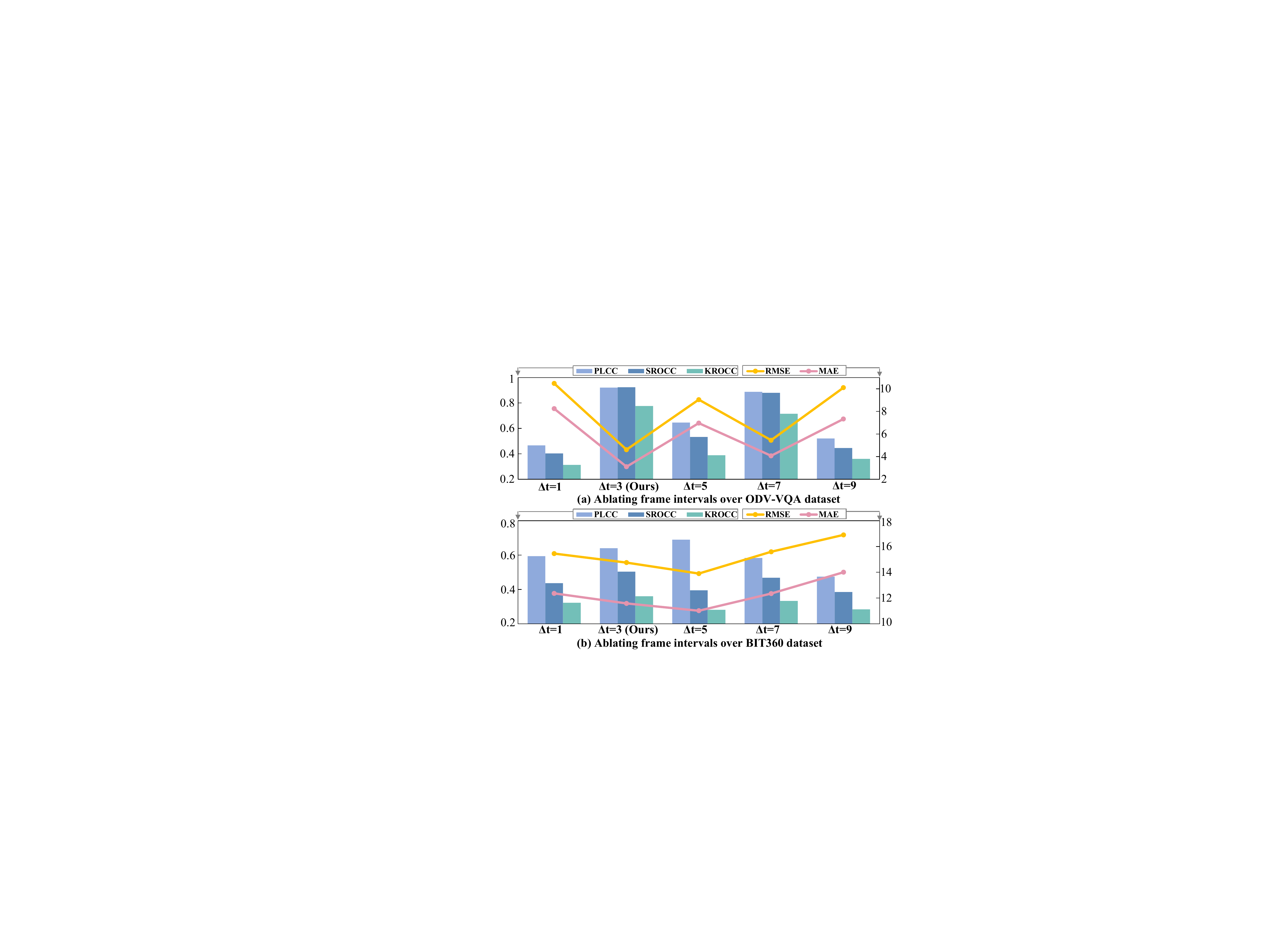}
	\end{center}
	\vspace{-1.em}
	\caption{Analysis on frame interval in our ProVQA approach.}
	\label{abl_FI}
	\vspace{-1.6em}
\end{figure}

\textbf{Ablation on the MFTN sub-net.}
In our approach, the developed MFTN sub-net is essential for capturing the contextual quality correlation among inconsecutive frames.   
To evaluate its effectiveness, we substitute the MFTN sub-net by four common frameworks of temporal prediction, \textit{i.e.}, long short-term memory (LSTM), convolutional LSTM (ConvLSTM), RNN and C3D.
The same as MFTN, both ConvLSTM and C3D take the frame-level quality maps as input, while both LSTM and RNN take the vectors flattened from the frame-level quality maps as input. 
Our ablation experiments are also conducted over the datasets of ODV-VQA and BIT360, and the results are presented in Table \ref{table:abl_5}. From this table, we can see that MFTN beats all other frameworks over ODV-VQA dataset with at least 0.014 increment in PLCC and 0.022 increment in SROCC.
Similar improvement can be found for the BIT360 dataset.
This validates the positive effect of the proposed MFTN sub-net to our ProVQA approach.

\textbf{Analysis on hyper-parameter $\Delta t$.}
In our experiment, $\Delta t$ denotes the frame interval between adjacent frames, which determines the estimated motion maps. With respect to the frame interval, a small value of $\Delta t$ may cause little motion information utilized by our approach, while a large one may introduce improper noise in the motion maps. Thus, a proper setting of frame interval $\Delta t$ is crucial for BVQA on 360{\textdegree} video. We thus evaluate the effects of $\Delta t$ over two datasets, by setting $\Delta t$ to 1, 3, 5, 7 and 9, respectively. We can see from Figure \ref{abl_FI} that $\Delta t$=3 offers the best performance, and therefore it is also the default setting in our approach. 
     
\textbf{Analysis on hyper-parameter $S$.}  
In our ProVQA approach, the hyper-parameter $S$ denotes the number of sampled 360{\textdegree} video clips. 
In our experiment, $S$ is maximally set to 6, due to the constraint of GPU memory. Here, we investigate whether the number of video clip samples influences the BVQA performance at $S$=1, 2, 3, 4 and 5. Note that when $S$ is set to 1, our MFTN sub-net is relaxed to the naive spatial non-local structure. The results of our ProVQA approach with different values of $S$ are shown in Table \ref{table:abl_6}. From this table, we can see that the setting of $S$=6 can obtain the best performance, while the setting of $S$=1 has the worst performance. This indicates that along with the increased number of sampled video clips, our approach can capture more long-term information, thus benefiting for BVQA on 360{\textdegree} video.

\begin{table}[!tb]
	\renewcommand\arraystretch{1.28}
	\centering
	\caption{Analysis on number of video clip samples, over VQA-ODA and BIT360 datasets.}
	\vspace{.8em}
	\resizebox{1\linewidth}{!}{
		\footnotesize
		\begin{tabular}{|c|cccccc|}
			\hline
			\multirow{1}{*}{\textbf{Metrics}} & \multicolumn{6}{c|}{\textbf{Ablation on VQA-ODV dataset}} \\
			\hline
			& \multicolumn{1}{c|}{$S$=6 (ours)} & $S$=1 & $S$=2 & $S$=3 & $S$=4 & $S$=5  \\ 
			\cline{2-7}
			\multirow{1}{*}{PLCC $\uparrow$} & \multicolumn{1}{c|}{\textbf{\color{red}{0.9209}}} &
			0.3050 & 0.3191  & \color{green}{\underline{0.5240}}  & \color{cyan}{\underline{0.8348}}  & 0.4677  \\
			\multirow{1}{*}{SROCC $\uparrow$} & \multicolumn{1}{c|}{\textbf{\color{red}{0.9236}}} & 0.1551  & 0.3509 & \color{green}{\underline{0.5222}} & \color{cyan}{\underline{0.8354}}  & 0.4093  \\
			\multirow{1}{*}{KROCC $\uparrow$} & \multicolumn{1}{c|}{\textbf{\color{red}{0.7760}}}  & 0.0897  & 0.2458  & \color{green}{\underline{0.4300}}  & \color{cyan}{\underline{0.6677}} &  0.3018 \\
			\multirow{1}{*}{RMSE $\downarrow$} & \multicolumn{1}{c|}{\textbf{\color{red}{4.6165}}} & 11.2808  & 11.2258  & \color{green}{\underline{10.0885}} & \color{cyan}{\underline{6.5206}} & 10.4695  \\
			\multirow{1}{*}{MAE $\downarrow$} & \multicolumn{1}{c|}{\textbf{\color{red}{3.1136}}} & 9.3233 & 9.1508 & \color{green}{\underline{7.6601}} & \color{cyan}{\underline{4.0931}}  & 8.8117  \\
			\hline
			\multirow{1}{*}{\textbf{Metrics}} & \multicolumn{6}{c|}{\textbf{Ablation on BIT360 dataset}} \\
			\hline
			& \multicolumn{1}{c|}{$S$=6 (ours)} & $S$=1 & $S$=2 & $S$=3 & $S$=4 & $S$=5  \\
			\cline{2-7}
			\multirow{1}{*}{PLCC $\uparrow$} & \multicolumn{1}{c|}{\textbf{\color{red}{0.6399}}} & 0.3550  & 0.5073 & 0.3806  & \color{green}{\underline{0.5306}}  & \color{cyan}{\underline{0.5818}}\\
			\multirow{1}{*}{SROCC $\uparrow$} & \multicolumn{1}{c|}{\textbf{\color{red}{0.5038}}} & 0.2257 & \color{green}{\underline{0.4411}}  & 0.3315 & \color{cyan}{\underline{0.4764}}  & 0.3028 \\
			\multirow{1}{*}{KROCC $\uparrow$} & \multicolumn{1}{c|}{\textbf{\color{red}{0.3605}}} & 0.1602 & \color{green}{\underline{0.2935}} & 0.2667 & \color{cyan}{\underline{0.3234}} & 0.2397 \\
			\multirow{1}{*}{RMSE $\downarrow$} & \multicolumn{1}{c|}{\textbf{\color{red}{14.7513}}}  & 17.9464  & 16.5424  & 17.7517  & \color{green}{\underline{16.2708}}  &  \color{cyan}{\underline{15.6126}} \\
			\multirow{1}{*}{MAE $\downarrow$} & \multicolumn{1}{c|}{\textbf{\color{red}{11.5803}}} & 14.6453  & 12.9510 & 14.9474 & \color{green}{\underline{12.7928}} & \color{cyan}{\underline{12.3300}} \\
			\hline		
		\end{tabular}
	}
	\label{table:abl_6}
	\vspace{-1.6em}
\end{table}

\vspace{-0.7em}

\section{Conclusion}

In this paper, we have proposed a novel BVQA approach for 360{\textdegree} video, called ProVQA.
To accord with the progressive perception mechanism of human, we designed the sub-nets of SPAQ, MPAQ and MFTN in our ProVQA approach, which learn from pixels, frames and video, respectively, for modeling and assessing the quality degradation of 360{\textdegree} video. 
First, by learning from pixels, the proposed SPAQ sub-net models the spatial quality degradation at each video frame, inspired by spherical perception mechanism of human. Then, by learning from frames, the MPAQ sub-net learns the motion artifacts through the motion cues across adjacent frames. Finally, by learning from video, the developed MFTN sub-net aggregates multi-frame quality degradation to yield the final quality score, via exploring long-term quality correlation of 360{\textdegree} video. Extensive experimental results showed the superiority and high generalization ability of our ProVQA approach by comparing with other 13 state-of-the-art approaches.

There are two promising directions for the future works.
(1) Since it is extremely time-consuming to establish a large-scale VQA dataset, the unsupervised or weakly-supervised BVQA approaches for 360{\textdegree} video are urgently needed in the future.
In particular, the few-shot learning paradigm may be embedded in our ProVQA approach for weakly-supervised BVQA on 360{\textdegree} video.
(2) The potential applications of our BVQA approach can be seen as another future direction. 
For example, the BVQA metrics by our approach can be used to guide the optimization of 360{\textdegree} video compression, such that better QoE can be achieved when viewing 360{\textdegree} video.


\ifCLASSOPTIONcaptionsoff
\newpage
\fi

\bibliographystyle{IEEEtran}
\bibliography{IEEEexample}

\vfill

\end{document}